\documentclass[10pt,twocolumn,letterpaper]{article}

\usepackage[pagenumbers]{cvpr} 

\newcommand{\rec}[1]{#1}   
\newcommand{\dtime}{\mathsf{t}}   
\newcommand{\sd}[1]{{\scriptsize$\pm$#1}}
\newcommand{\dlt}[1]{{\scriptsize(#1)}}
\newcommand{\cellsd}[3]{\begin{tabular}{@{}c@{}}#1\sd{#2}\\[-2pt]\dlt{#3}\end{tabular}}

\newcommand{\cellms}[3]{\begin{tabular}{@{}c@{}}#1\sd{#2}\\[-2pt]\dlt{#3}\end{tabular}}

\usepackage[utf8]{inputenc} 
\usepackage[T1]{fontenc}    
\usepackage{url}            
\usepackage{booktabs}       
\usepackage{amsfonts}       
\usepackage{nicefrac}       
\usepackage{microtype}      
\usepackage{xcolor}         
\usepackage{pifont}         

\usepackage{algorithm}
\usepackage{algpseudocode}

\usepackage{amsmath}
\usepackage{amssymb}
\usepackage{amsfonts}
\usepackage{bbm}
\usepackage{mathtools}
\usepackage{amsthm}
\usepackage{mathrsfs}
\usepackage{tabularx}
\usepackage{titletoc}
\usepackage{appendix}
\usepackage{multirow}
\usepackage{booktabs}
\usepackage{subcaption}
\usepackage{graphicx}
\usepackage{wrapfig}

\definecolor{cvprblue}{rgb}{0.21,0.49,0.74}
\definecolor{degrade}{RGB}{205,80,80}   
\definecolor{improve}{RGB}{50,110,200}   
\usepackage[pagebackref,breaklinks,colorlinks,allcolors=cvprblue]{hyperref}

\usepackage[capitalize,noabbrev,nameinlink]{cleveref}
\newcommand{\crefp}[2]{\hyperref[#1]{\cref*{#1}#2}}
\newcommand{\Crefp}[2]{\hyperref[#1]{\Cref*{#1}#2}}
\crefname{figure}{Fig.}{Figs.}
\Crefname{figure}{Figure}{Figures}
\crefname{equation}{Eq.}{Eqs.}
\Crefname{equation}{Equation}{Equations}
\newcommand{\methodname}{PlanNudge\xspace}

\def\confName{CVPR}
\def\confYear{2027}

\title{Your Diffusion-Based Driving Planner Only Needs a Nudge}

\author{
Zilin Wang$^{1*}$ \quad Guozheng Ma$^{2}$ \quad Saeed Rahmani$^{3}$ \quad Jiong Yang$^{4}$\\
Antyanta Bangunharcana$^{4}$ \quad Yiluan Guo$^{4\dagger}$ \quad Jakob Nicolaus Foerster$^{1}$ \quad Shimon Whiteson$^{1}$\\[4pt]
$^{1}$BOLD, University of Oxford \quad $^{2}$Nanyang Technological University\\
$^{3}$Delft University of Technology \quad $^{4}$Motional
}

\begin{document}
\makeatletter
\twocolumn[{
  \@maketitle
  \begin{center}
    \setlength{\abovecaptionskip}{3pt}
    \vspace*{-17pt}
    \includegraphics[width=\textwidth]{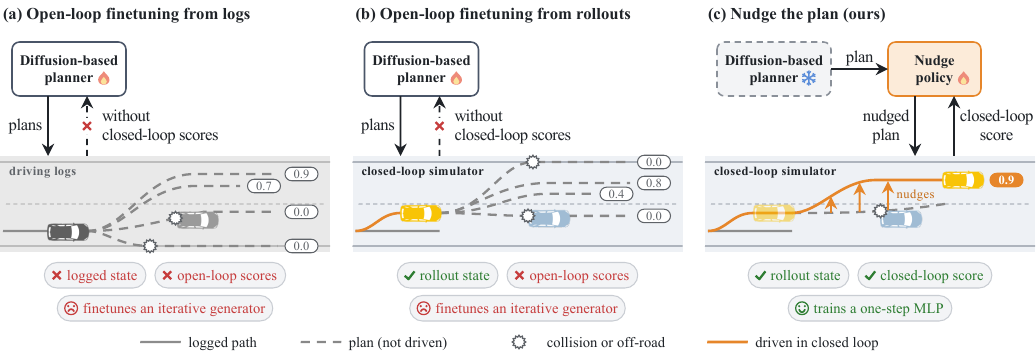}
    \captionof{figure}{
    \textbf{Reinforcement learning for post-training diffusion-based planners.}
    (a)~Open-loop finetuning from logs updates the planner by scoring candidate plans sampled from a logged state without executing them.
    (b)~Methods such as PlannerRFT~\cite{li2026plannerrft} sample candidate plans at states visited during closed-loop rollouts but still update the planner using open-loop scores.
    (c)~We freeze the diffusion-based planner and use a one-step MLP to nudge its plan.
    We execute the nudged plan in simulation and train the MLP using the resulting closed-loop score.
    }
    \label{fig:teaser}
    \vspace*{-1pt}
    {\setlength{\unitlength}{\textwidth}
    \begin{picture}(1,0.1706)   
      \put(0,0){\includegraphics[width=\textwidth]{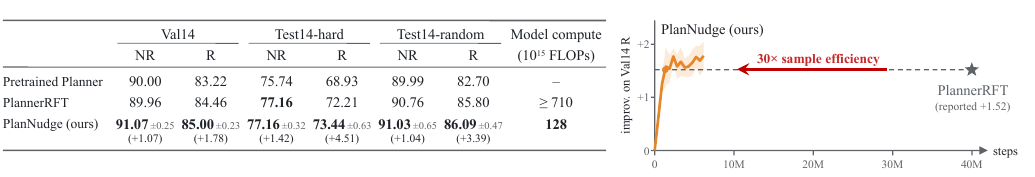}}
\put(0.0709,0.0719){\makebox(0,0)[l]{\fontsize{6.5}{8}\selectfont\citep{li2026plannerrft}}}

    \end{picture}}
    \captionof{figure}{ 
    \textbf{Results on nuPlan.}
    \methodname matches or exceeds the state-of-the-art PlannerRFT in all six settings with about one sixth of its environment steps and less than one fifth of its model compute.
    In our simulator, it reaches PlannerRFT's reported Val14 R gain (+1.52 over its own score of the pretrained planner) after about 1.3\,M steps, one thirtieth of 40\,M.
    Mean and standard deviation over six runs.
    }
    \label{fig:results}
    \vspace*{2pt}
  \end{center}
}]
\makeatother
{\renewcommand{\thefootnote}{}
\footnotetext{$^{*}$Work done during an internship at Motional.}%
\footnotetext{$^{\dagger}$Corresponding author: {\tt yiluan.guo@motional.com}}}
\begin{abstract}
Diffusion-based planners capture the multimodality of human driving behavior well, yet covariate shift in closed-loop driving can lead them to generate trajectories that result in collisions or road departures.
Reinforcement learning (RL) finetuning offers a promising way to tackle this issue, but applying it to diffusion-based planners is complicated by their iterative denoising process and receding-horizon execution.
We find that small adjustments to the planned trajectories can prevent many of these unsafe outcomes.
Motivated by this observation, we propose \methodname, which preserves the learned planning capabilities of a pretrained diffusion-based planner by freezing it and trains a lightweight policy that we call the nudge policy.
At each planning step, the nudge policy outputs a nudge, a two-dimensional adjustment of the speed and curvature of the plan from the frozen planner.
The nudge policy is a multilayer perceptron that predicts the nudge in a single forward pass, and both dimensions directly change the motion of the vehicle, so standard off-policy RL can train it in closed-loop simulation.
We evaluate \methodname on nuPlan, a large-scale autonomous driving benchmark based on real-world driving data.
In all six settings, \methodname matches or exceeds the final scores of the previous state-of-the-art RL finetuning method while using about one sixth as many environment steps and less than one fifth as much model compute.
We also provide guidelines for reward shaping, value learning, and exploration when using off-policy RL to post-train autonomous driving planners.
Our results are reported with means and standard deviations over six training runs.
Project page: \url{https://zerlinwang.github.io/plannudge/}.
\end{abstract}
    
\section{Introduction}
\label{sec:intro}



Trajectory planning in autonomous driving requires anticipating how other vehicles will move and deciding when to proceed or yield~\cite{rhinehart2019precog}.
Human demonstrations of these interactions are often multimodal~\cite{zheng2025diffusionplanner,zheng2026hdp}.
A planner trained to predict a single trajectory with a mean squared error loss can average those maneuvers into a plan that follows neither~\citep{cui2019multimodal}.
Diffusion-based planners can represent these distinct maneuvers by learning a distribution over trajectories conditioned on the traffic scene~\citep{zheng2025diffusionplanner,liao2025diffusiondrive}.
They perform well in simulation~\cite{zheng2025diffusionplanner,tan2025flowplanner} and in real-world driving~\cite{zheng2026hdp}.

Despite their success, diffusion-based planners can still collide with other vehicles or leave the road in closed-loop driving~\citep{zheng2025diffusionplanner,seyam2026model}.
One contributing factor is covariate shift: a planner trained on logged demonstrations can encounter unfamiliar states as its own errors accumulate~\citep{ross2011dagger}.
Reinforcement learning (RL) post-training can help address this mismatch by allowing the planner to learn from the states it visits in closed-loop simulation, using rewards that reflect the consequences of its driven trajectories~\citep{tang2025planr1,zhang2025carplanner,cusumano-towner2025robust,wang2026learning}.


However, applying RL to diffusion-based planners raises two difficulties.
The first difficulty comes from iterative denoising: a plan comes from several denoising steps, so its likelihood is intractable.
Policy-based methods therefore treat each denoising step as an action \citep{ren2025dppo,liu2025flowgrpo}, which multiplies the horizon of the decision process by the number of denoising steps~\citep{psenka2024qsm,ren2025dppo,chen2025pirl}.
Value-based methods update policies by backpropagating the critic’s gradient through all denoising steps, which is costly and can destabilize training \citep{wang2023diffusionql,kang2023edp,park2025fql}.
The second difficulty arises from receding-horizon execution.
To remain responsive to changing traffic, a driving planner may generate a trajectory spanning several seconds but execute only a single control interval before replanning~\cite{karnchanachari2024nuplan}.
When the full trajectory is treated as the RL action, the environment provides feedback on the executed portion while the rest of the plan is superseded.
This mismatch complicates value learning~\citep{ren2025dppo,chen2025pirl}.

The first difficulty can be addressed by casting iterative denoising as a Markov decision process, which enables policy gradient training at the cost of a longer RL horizon \citep{black2024ddpo,ren2025dppo}.
ReCogDrive~\citep{li2025recogdrive} and DiffusionDriveV2~\citep{zou2025diffusiondrivev2} use this formulation to finetune their planners on open-loop scores of plans sampled in logged states (\crefp{fig:teaser}{a}).
PlannerRFT~\citep{li2026plannerrft} addresses the resulting state distribution gap by training on states visited during closed-loop rollouts.
Its exploration policy learns from rollout rewards, while its diffusion-based planner learns from open-loop plan scores (\crefp{fig:teaser}{b}).
Scoring whole plans in open loop avoids the second difficulty, at the cost of a gap to the closed-loop score of the driven trajectory, since the tracker follows only a prefix of each plan~\cite{dauner2023parting}.
It remains open whether RL can improve diffusion-based planners from closed-loop scores efficiently.

All these methods finetune the diffusion-based planner itself, which is where both difficulties arise.
This raises a natural question: \textit{can we post-train diffusion-based planners without finetuning them?}
As a first test, we keep the planner fixed and add small random adjustments to the speed and curvature of its plans (\cref{sec:analysis}).
On nuPlan~\cite{karnchanachari2024nuplan}, at least one of 20 closed-loop trials avoids the failure in more than half of the scenarios that a pretrained Diffusion Planner~\cite{zheng2025diffusionplanner} fails, while sampling the planner again avoids almost none.


Based on this observation, we propose \methodname (\crefp{fig:teaser}{c}), which keeps the diffusion-based planner frozen and learns the adjustment with a nudge policy.
At each planning step, the nudge policy outputs a two-dimensional nudge, a speed offset and a curvature offset, to adjust the plan.
This design addresses both difficulties.
For iterative denoising, the diffusion-based planner is frozen and the nudge is generated in one forward pass, so the RL agent makes one decision per plan, whatever the number of denoising steps.
For receding-horizon execution, the action space is small and both action dimensions change the executed portion of the plan.
The critic assigns credit to two numbers that matter instead of to a whole plan of which only the beginning is driven.
Freezing the planner also preserves its pretrained multimodal distribution over maneuvers, and a zero nudge recovers the sampled plan exactly.
We train the nudge policy on closed-loop returns with off-policy RL.


We evaluate \methodname on nuPlan~\cite{karnchanachari2024nuplan}, a closed-loop planning benchmark built on large-scale real driving logs.
Compared with the state-of-the-art RL finetuning method PlannerRFT~\cite{li2026plannerrft}, starting from the same pretrained Diffusion Planner~\cite{zheng2025diffusionplanner}, \methodname matches or exceeds its reported scores in all six settings with 6.2\,M environment steps, about one sixth of PlannerRFT's reported 40\,M, and with less than one fifth of its model compute, the floating-point operations of all network passes during RL training (\cref{sec:supp_flops}).
We also demonstrate the effectiveness of \methodname on Hyper Diffusion Planner (HDP)~\citep{zheng2026hdp} with both reactive and non-reactive training traffic modes.
Our results are means and standard deviations over six training runs.
Our main contributions are:
\begin{itemize}
    \item We propose \methodname, which freezes the diffusion-based planner and trains a nudge policy with off-policy RL in closed loop.
    The nudge policy generates its action in a single forward pass, without iterative denoising, so the RL horizon stays short.
    The action has two dimensions, and both change the executed portion of the plan, which simplifies value learning under receding-horizon execution.
    \item We demonstrate on nuPlan that \methodname matches or exceeds the previous state-of-the-art RL finetuning method in all six settings, with about one sixth of its environment steps and less than one fifth of its model compute.
    \item We provide several guidelines for post-training autonomous driving planners with off-policy RL, covering reward shaping, value learning, and exploration.
\end{itemize}

\section{Related work}
\label{sec:related}

\paragraph{Diffusion-based driving planners.}
Diffusion models~\citep{ho2020ddpm} and flow matching~\citep{lipman2023flow} are widely used to generate robot actions~\citep{chi2023diffusionpolicy} and vehicle trajectories~\citep{zheng2025diffusionplanner}.
Diffusion Planner~\citep{zheng2025diffusionplanner} generates the ego plan together with the future trajectories of the surrounding traffic participants, and Flow Planner~\citep{tan2025flowplanner} uses flow matching to model interactive behavior.
Both achieve strong closed-loop results on nuPlan.
DiffusionDrive~\citep{liao2025diffusiondrive} denoises from anchor trajectories with a truncated schedule for end-to-end driving, and Hyper Diffusion Planner~\citep{zheng2026hdp} scales diffusion-based planning to large real-world datasets and real-vehicle tests.
These planners learn from logged demonstrations by imitation, which brings the known limitations of behavior cloning~\citep{codevilla2019limitations}.
The imitation loss carries no notion of driving quality, so a plan that nearly hits another vehicle costs no more than one that keeps clear.
The planner is trained only in logged states, so its own errors lead it into states that the logs never cover, where the errors compound (covariate shift)~\citep{ross2011dagger}.
It can also rely on spurious cues in the scene instead of the causes of the demonstrated behavior (causal confusion)~\citep{dehaan2019causal,sun2024causalagents}.
Plans that look right in open loop therefore still end in collisions or off the road in closed loop~\citep{dauner2023parting,zheng2025diffusionplanner,seyam2026model}.
Data augmentation and auxiliary losses mitigate these limitations~\citep{cheng2024pluto}, but they cannot cover the states that the planner itself produces, which are only visited by closed-loop training.

\vspace{-0.5\baselineskip}
\paragraph{RL post-training for diffusion-based driving planners.}
RL post-training supplies the driving objective that imitation lacks: the planner is rewarded for the quality of its plans instead of their closeness to the log~\citep{huang2025gendrive,li2025recogdrive,zou2025diffusiondrivev2,li2026plannerrft}.
Iterative denoising makes the likelihood of a plan intractable, so RL methods for diffusion models treat the denoising process as a Markov decision process~\citep{black2024ddpo,ren2025dppo,liu2025flowgrpo}, backpropagate a critic's gradient through the denoising steps~\citep{wang2023diffusionql}, or weight the denoising objective by a reward~\citep{zheng2026hdp}.
Most driving methods apply them in logged states: they sample several plans per state and move the planner toward the plans that a learned reward model~\citep{huang2025gendrive}, human feedback~\citep{li2025trajhf}, or rule-based trajectory scores~\citep{li2025recogdrive,zheng2026hdp,liu2025r2se} rank highest.
Two gaps remain: the planner still never visits the states that its own driving produces (the state gap), and the score of a plan is not the outcome of executing its beginning and replanning (the reward gap).
PlannerRFT~\citep{li2026plannerrft} closes the state gap: an exploration policy trained in closed loop with PPO~\cite{schulman2017ppo} guides the sampling, so the planner is updated in the states of its own rollouts.
Its reward remains the open-loop score of a sampled plan, its planner is updated with policy gradient~\cite{black2024ddpo} and group-relative advantage~\citep{shao2024grpo}, and its two policies are trained on-policy, which takes 40\,M environment steps in its reported setup.
\methodname closes both gaps with one policy: the reward is the closed-loop return of the driven trajectory, and a nudge policy learns from it with off-policy RL while the planner stays frozen, so RL makes one decision per planning step, not per denoising step.

\vspace{-0.5\baselineskip}
\paragraph{Learning residual policies.}
Residual policy learning improves a fixed controller with a learned correction~\citep{silver2018residual,johannink2019residual}.
Recent work trains such a residual on top of a pretrained robot policy with on-policy~\citep{ankile2024resip} or off-policy~\citep{yuan2025policydecorator,ankile2026resfit,wagenmaker2025steering,ma2026what} RL.
A small residual policy brings large gains in manipulation.
Motivated by the effect of small random adjustments (\cref{sec:analysis}), \methodname applies it to closed-loop trajectory planning for the first time, to our knowledge.
This application raises two challenges.
First, a plan is a long sequence of waypoints, which is a large action space for a residual.
\methodname therefore nudges only the speed and curvature of the plan, a two-dimensional action that changes the executed portion of the plan.
Second, the downstream tracker filters out small per-step changes of the plan, so exploration uses temporally correlated noise and scripted speed events to reach the maneuvers that avoid failures.
As the score of a drive arrives only at the end of an episode, \methodname also learns values with multi-step returns.
Without multi-step returns or correlated noise, much of the gain is lost (\cref{sec:exp_takeaways}).
In addition, a dense safety penalty shapes the reward.

\section{Preliminaries}
\label{sec:prelim}

\subsection{Trajectory planning and diffusion models}
\label{sec:prelim_planning}

\begin{figure*}[t]
    \centering
    \includegraphics[width=\textwidth]{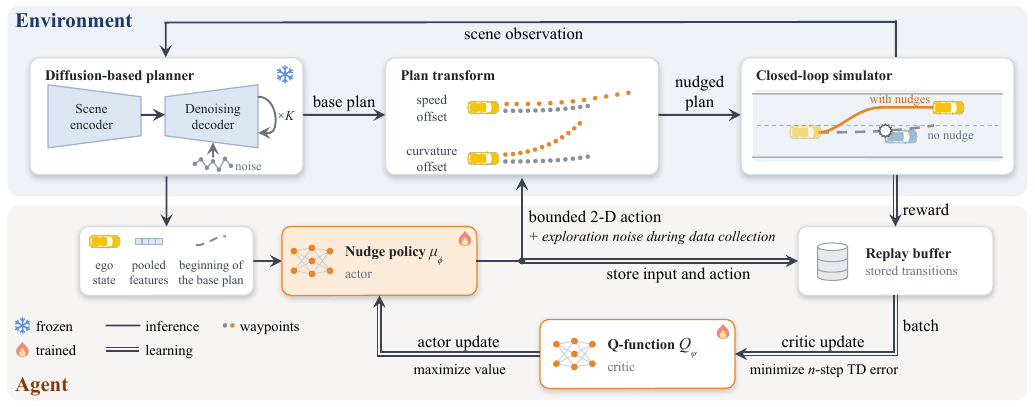}
    \vspace{-1.5\baselineskip}
    \caption{\textbf{Overview of \methodname.}
    Starting from random noise, the frozen pretrained diffusion-based planner samples a base plan with $K$ denoising steps.
    The nudge policy reads the ego state, the pooled scene features, and the beginning of the base plan, and outputs a bounded 2-D action, which the plan transform converts into a speed offset and a curvature offset of the base plan.
    The simulator follows the nudged plan for one replanning interval $\Delta t$ before it returns the next scene observation and the planner replans.
    Each transition is stored in a replay buffer.
    The diffusion-based planner is a part of the \textit{environment} and is never called during RL \textit{agent} updates.
    }
    \label{fig:method}
    \vspace{-0.5\baselineskip}
\end{figure*}

We consider trajectory planning as a closed-loop receding-horizon procedure~\citep{karnchanachari2024nuplan}.
At each planning step $t$, the planner receives a scene observation $o_t$, containing information about the ego vehicle, the surrounding traffic participants, and the map.
It produces a plan $\tau_t = (w_t^1,\dots,w_t^H)$ of $H$ future waypoints, each specifying a position and heading.
A tracker converts the plan into control commands, which are applied for an interval $\Delta t$ before the planner receives $o_{t+1}$ and replans.
The planning horizon typically exceeds $\Delta t$, so each plan is replaced before it is fully executed.
Closed-loop performance is therefore evaluated on the driven trajectory, which results from tracking and successive replanning.

A diffusion-based planner learns a conditional distribution over trajectories from logged demonstrations.
At each planning step $t$, it generates a trajectory by progressively transforming Gaussian noise into a plan conditioned on $o_t$.
For ODE-based sampling, this reverse generation process follows the probability-flow ODE~\citep{song2021sde},
\begin{equation}
    \frac{\mathrm{d}\mathbf{x}_{\dtime}}{\mathrm{d}\dtime}
    =
    f(\dtime)\mathbf{x}_{\dtime}
    - \frac{1}{2}g(\dtime)^2
    \nabla_{\mathbf{x}}\log p_{\dtime}(\mathbf{x}_{\dtime}\mid o_t),
    \qquad \dtime:1\rightarrow 0,
    \label{eq:pfode}
\end{equation}
initialized with $\mathbf{x}_1 \sim \mathcal{N}(\mathbf{0},\mathbf{I})$.
Here, $\dtime$ denotes the diffusion time, $f$ and $g$ are specified by the noise schedule, and the conditional score $\nabla_{\mathbf{x}}\log p_{\dtime}(\mathbf{x}_{\dtime}\mid o_t)$ is estimated by the learned denoising network $\epsilon_\theta(\mathbf{x}_{\dtime},\dtime,o_t)$ up to the scaling of the noise schedule.
A numerical solver integrates this ODE in $K$ steps, with one forward pass per step~\citep{song2021ddim, lu2025dpmsolverpp}.

\vspace{-0.5\baselineskip}
\subsection{Off-policy reinforcement learning}
\label{sec:prelim_rl}

We formulate closed-loop planning as a Markov decision process $(\mathcal{S},\mathcal{A},P,r,\gamma)$ whose state $s_t \in \mathcal{S}$ is the scene observation $o_t$, which we assume to be a Markov state.
At each planning step $t$, which is one environment step, a policy $\pi(a_t \mid s_t)$ selects an action $a_t \in \mathcal{A}$ that determines the plan supplied to the tracker.
The transition function $P(s_{t+1}\mid s_t,a_t)$ incorporates plan construction, tracking, vehicle dynamics, and the motion of the surrounding traffic participants.
The reward $r_t = r(s_t,a_t,s_{t+1})$ evaluates the resulting scene.
The objective is to maximize the expected discounted return $     J(\pi) =
     \mathbb{E}_{\pi}
     \left[\sum_{t=0}^{\infty}\gamma^t r_t\right]$.

The action-value function $Q^\pi(s,a)$ is the expected discounted return conditioned on taking action $a$ in state $s$ and following $\pi$ thereafter.
Off-policy actor-critic methods learn from transitions $(s_t,a_t,r_t,s_{t+1})$ stored in a replay buffer $\mathcal{B}$.
DDPG~\citep{lillicrap2016ddpg} and TD3~\citep{fujimoto2018td3} learn a critic $Q_\psi$ through temporal-difference updates and optimize a deterministic actor $\mu_\phi$ using the surrogate objective
$J_{\mathrm{actor}}(\phi)=\mathbb{E}_{s\sim\mathcal{B}}[Q_\psi(s,\mu_\phi(s))]$.
Its gradient is:
\begin{equation}
    \nabla_\phi J_{\mathrm{actor}}
    =
    \mathbb{E}_{s\sim\mathcal{B}}
    \left[
        \nabla_\phi \mu_\phi(s)\,
        \left.\nabla_a Q_\psi(s,a)\right|_{a=\mu_\phi(s)}
    \right].
    \label{eq:actor}
\end{equation}
Experience replay enables repeated optimization using previously collected interactions, which improves sample efficiency.
Unlike on-policy RL, which can only learn from actions sampled by the current policy, off-policy RL also admits transitions whose actions are not from the actor: a scripted maneuver executed during data collection enters the critic update like any other transition, and the actor adopts it where the critic values it~\cite{vecerik2017leveraging, ball2023efficient}.
This lets \methodname inject the maneuvers that avoid driving failures (\cref{sec:exp_takeaways}).

\section{PlanNudge}
\label{sec:method}

\vspace{-0.2\baselineskip}
\subsection{Nudging a frozen planner}
\label{sec:method_system}

\begin{table*}[t]
\centering
\small
\setlength{\tabcolsep}{3.35pt}
\renewcommand{\arraystretch}{1.15}
\caption{
\textbf{Closed-loop scores on nuPlan, official devkit.}
Starting from the Diffusion Planner (DDIM), open-loop finetuning from logs and \methodname are implemented by us and reported as the mean and standard deviation over six runs, with the change in brackets.
Bold denotes the highest result in each setting.
Red: worse than the pretrained planner.
Blue: an improvement of more than 1 point.
}
\vspace{-0.8\baselineskip}
\begin{tabular}{@{}llccccccc@{}}
\toprule
& & \multicolumn{2}{c}{Val14} & \multicolumn{2}{c}{Test14-hard} & \multicolumn{2}{c}{Test14-random} & RL \\
\cmidrule(lr){3-4} \cmidrule(lr){5-6} \cmidrule(lr){7-8}
Type & Method & NR & R & NR & R & NR & R & steps \\
\midrule
\textcolor{gray}{Expert} & \textcolor{gray}{Log replay} & \textcolor{gray}{93.53} & \textcolor{gray}{80.32} & \textcolor{gray}{85.96} & \textcolor{gray}{68.80} & \textcolor{gray}{94.03} & \textcolor{gray}{75.86} & \textcolor{gray}{--} \\
\midrule
\multirow{5}{*}{Imitation learning}
& PLUTO~\citep{cheng2024pluto} & 88.89 & 78.11 & 70.03 & 59.74 & 89.90 & 78.62 & -- \\
& Diffusion Planner~\citep{zheng2025diffusionplanner} & 89.87 & 82.80 & 75.99 & 69.22 & 89.19 & 82.93 & -- \\
& Flow Planner~\citep{tan2025flowplanner} & 90.43 & 83.31 & 76.47 & 70.42 & 89.88 & 82.93 & -- \\
& DFP~\citep{zhang2026dfp} & 90.33 & 79.97 & 76.91 & 63.56 & 90.69 & 81.96 & -- \\
& Diffusion Planner (DDIM) & \rec{90.00} & \rec{83.22} & \rec{75.74} & \rec{68.93} & \rec{89.99} & \rec{82.70} & -- \\
\midrule
\multirow{3}{*}{RL post-training}
& Open-loop finetuning from logs & \cellms{\rec{90.09}}{\rec{0.16}}{\rec{+0.09}} & \cellms{\rec{84.35}}{\rec{0.20}}{\textcolor{improve}{\rec{+1.13}}} & \cellms{\rec{76.39}}{\rec{0.89}}{\rec{+0.65}} & \cellms{\rec{70.42}}{\rec{0.20}}{\textcolor{improve}{\rec{+1.49}}} & \cellms{\rec{89.95}}{\rec{0.43}}{\textcolor{degrade}{\rec{\textminus 0.04}}} & \cellms{\rec{85.02}}{\rec{0.23}}{\textcolor{improve}{\rec{+2.32}}} & -- \\
& PlannerRFT~\citep{li2026plannerrft} & 89.96 & 84.46 & \textbf{77.16} & 72.21 & 90.76 & 85.80 & 40\,M \\
& \methodname (ours) & \cellms{\textbf{\rec{91.07}}}{\rec{0.25}}{\textcolor{improve}{\rec{+1.07}}} & \cellms{\textbf{\rec{85.00}}}{\rec{0.23}}{\textcolor{improve}{\rec{+1.78}}} & \cellms{\textbf{\rec{77.16}}}{\rec{0.32}}{\textcolor{improve}{\rec{+1.42}}} & \cellms{\textbf{\rec{73.44}}}{\rec{0.63}}{\textcolor{improve}{\rec{+4.51}}} & \cellms{\textbf{\rec{91.03}}}{\rec{0.65}}{\textcolor{improve}{\rec{+1.04}}} & \cellms{\textbf{\rec{86.09}}}{\rec{0.47}}{\textcolor{improve}{\rec{+3.39}}} & \rec{6.2\,M} \\
\bottomrule
\end{tabular}
\label{tab:main}
\vspace{-1.0\baselineskip}
\end{table*}

As shown in \cref{fig:method}, \methodname uses a \textit{nudge policy} to nudge the plan from a frozen diffusion-based planner, which we call the \textit{base plan}, before feeding it to the closed-loop simulator.
A \textit{nudge} is a bounded two-dimensional action, and a fixed \textit{plan transform} turns it into a speed offset and a curvature offset of the base plan.
Specifically, at each planning step $t$, the frozen planner generates a base plan $\bar\tau_t = G_\theta(o_t,\mathbf{x}_1)$ from sampled isotropic Gaussian noise $\mathbf{x}_1$.
$G_\theta$ denotes the $K$-step numerical solution of the ODE in \cref{eq:pfode}.
The nudge policy is a multilayer perceptron (MLP) that observes $z_t=(f_t,e_t,\bar\tau_t^{0:T_{\mathrm{in}}})$.
Here, $f_t$ denotes the mean-pooled features of the pretrained frozen scene encoder.
The ego state $e_t$ is computed from the ego history in $o_t$ and holds the velocity, acceleration, steering angle, and yaw rate.
The input $\bar\tau_t^{0:T_{\mathrm{in}}}$ is the first $T_{\mathrm{in}}$ seconds of the base plan.
The policy then outputs the nudge in a single forward pass, and the fixed plan transform $T$ applies it to the base plan:
\begin{equation}
    a_t = \mu_\phi(z_t) = (a_t^{v},a_t^\kappa) \in [-1,1]^2,
    \qquad \tau_t = T(\bar\tau_t,a_t).
    \label{eq:nudge}
\end{equation}
The two components $(a_t^{v},a_t^\kappa)$ are the normalized speed offset and the normalized curvature offset.
$T$ ramps the speed offset up over a short initial window and then shifts the speed of the whole remaining plan, with braking clipped to prevent backward motion.
It bends the path with the curvature offset over a short initial window and then keeps the shape of the remaining plan.
A zero nudge returns the base plan exactly.
See \cref{sec:supp_transform} for details.

\subsection{Training the nudge policy}
\label{sec:method_training}

We train the nudge policy from closed-loop interaction using TD3~\citep{fujimoto2018td3}.
The nudge policy $\mu_\phi$ is the actor and a Q-function $Q_\psi$ is its critic, together forming the \textit{agent}.
The frozen planner, the transform $T$, and the simulator with its tracker form the \textit{environment} (\cref{fig:method}).
Each transition $(z_t,a_t,r_t,z_{t+1})$ is stored in a replay buffer $\mathcal{B}$.
Because the diffusion-based planner is part of the environment and $z_t$ already contains its outputs, updates never call any denoising process.
In the critic update, the critic $Q_\psi$ is updated with $n$-step temporal-difference learning.
Given the Q-function as the surrogate objective, the actor is updated by maximizing the Q-value while penalizing large nudges:
\begin{equation}
    \mathcal{L}(\phi)
    = \mathbb{E}_{z\sim\mathcal{B}}\Big[
    -Q_{\psi}\big(z,\mu_\phi(z)\big)
    +\lambda\big\|W\mu_\phi(z)\big\|_2^2\Big],
    \label{eq:actor_loss}
\end{equation}
where $W$ weights the two action components, and $\lambda$ controls the penalty.
In addition, the critic receives the remaining episode fraction $\rho_t$ as privileged information to account for the finite horizon, an asymmetric input that the actor does not see.
Both actor and critic are parameterized with SimBa~\citep{lee2025simba}.
REDQ~\citep{chen2021redq} is used to mitigate overestimation.
The actor's output layer is initialized to zero, so its initial deterministic action preserves the base plan.

We shape the reward by adding a small dense safety penalty to the terminal score of the nuPlan metrics.
The penalty grows as the time to collision, the lateral clearance, or the time to the road boundary shrinks (\cref{sec:supp_reward}).
For exploration, temporally correlated Ornstein--Uhlenbeck noise~\citep{lillicrap2016ddpg} is added to the nudge during data collection, so that the noise persists long enough to change the vehicle's motion through the tracker.
Additionally, scripted maneuvers, a brake or a push held for seconds, can be applied during exploration to supply the maneuvers that avoid driving failures.
We call them speed events.
For example, a push can be applied when the vehicle stays stopped although it should proceed, also known as the inertia problem~\cite{codevilla2019limitations}.

\vspace{-0.5\baselineskip}
\section{Experiments}
\label{sec:exp}

\vspace{-0.2\baselineskip}
\subsection{Experimental setup}
\label{sec:exp_setup}

\begin{figure*}[t]
    \centering
    \begin{subfigure}{\linewidth}
        \centering
        \includegraphics[width=\linewidth]{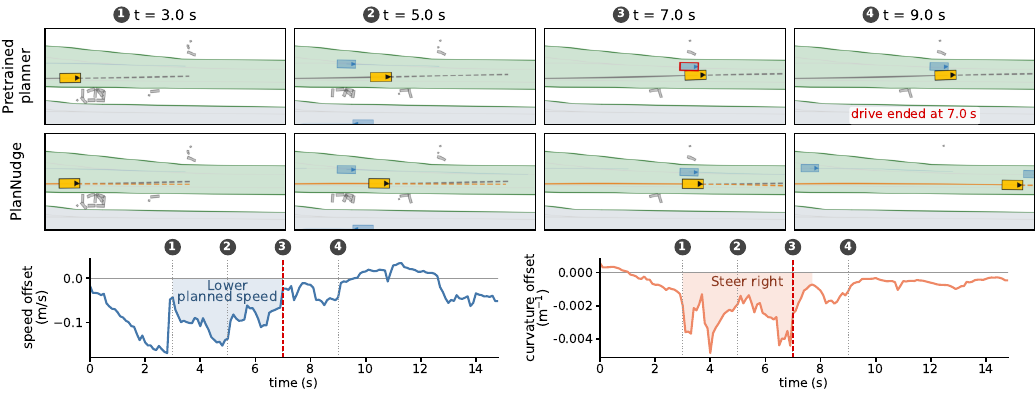}
        \caption{A faster vehicle approaches from behind. \methodname slows down and steers right, and the vehicle overtakes on the left.}
        \label{fig:qualitative_collision}
    \end{subfigure}
    \vspace{20pt}
    \begin{subfigure}{\linewidth}
        \centering
        \includegraphics[width=\linewidth]{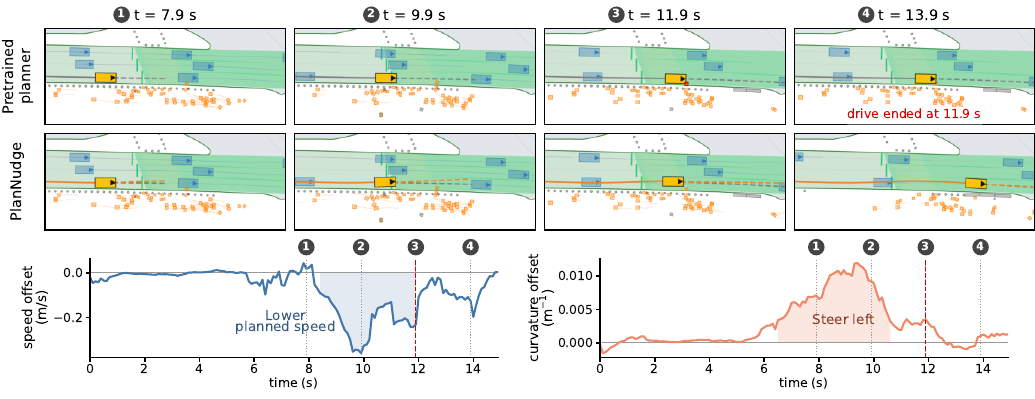}
        \caption{Pedestrians at the right edge of the lane. The pretrained planner hits one of them. \methodname slows down and steers left, away from them.}
        \vspace{-1.0\baselineskip}
        \label{fig:qualitative_pedestrians}
    \end{subfigure}
    \vspace{-2.0\baselineskip}
    \caption{
    \textbf{Qualitative results} under non-reactive (a) and reactive (b) traffic.
    Upper row: the pretrained planner, whose drive ends at the failure (red outline).
    Lower row: \methodname at the same times.
    Dashed: plans, drawn for their first 4\,s. Solid: driven paths.
    Gray: the pretrained planner. Orange: \methodname.
    The curves show the speed and curvature offsets. The numbers 1 to 4 mark the times of the four snapshots on the curves. The red dashed line marks the pretrained planner's failure. After it, the offsets return toward zero.
    }
    \label{fig:qualitative}
    \vspace{-0.5\baselineskip}
\end{figure*}

\paragraph{Benchmarks.}
We evaluate models on nuPlan~\citep{karnchanachari2024nuplan} with the official devkit.
nuPlan is a large-scale benchmark built on about 1,300 hours of real-world driving logs from four cities, and it is one of the most widely used closed-loop benchmarks for learned driving planners.
Three scenario sets are used.
Val14~\citep{dauner2023parting} draws 1,118 scenarios from the validation split.
Test14-hard~\cite{cheng2024rethinking} holds the 272 scenarios of the test split on which the rule-based planner PDM-Closed~\citep{dauner2023parting} scores lowest.
Test14-random~\citep{cheng2024rethinking} draws 261 scenarios from the test split at random.
Each scenario contains 20 frames of history, one current frame, and 150 frames of future trajectory at 10\,Hz, 171 frames in total.
Each set runs in two traffic modes, which gives six evaluation settings.
In non-reactive traffic (NR), the other traffic participants replay their logged trajectories.
In reactive traffic (R), the vehicles among them follow the Intelligent Driver Model (IDM)~\cite{Treiber_2000} and react to the ego vehicle and other traffic participants.
We name a setting by its set and its mode, for example Val14 R.
We report a mean closed-loop score between 0 and 100 for each setting.
Training, checkpoint selection, and the ablations of \cref{sec:exp_takeaways} run in a training environment built around the devkit's own simulation.
It runs nuPlan scenarios using the devkit’s LQR tracker, kinematic bicycle model, and reactive or non-reactive traffic.
It scores the driven trajectory at the end of each episode with a vectorized implementation of the devkit's metric definitions on cached map and scenario data, so many episodes run in parallel.
We call it ``our simulator''.

\vspace{-0.2\baselineskip}
\paragraph{Training details and baselines.}
We apply \methodname to the official checkpoint of the Diffusion Planner~\citep{zheng2025diffusionplanner}, a state-of-the-art diffusion-based planner on nuPlan, sampled with 5 DDIM steps and an initial noise scale of 0.1.
Our main baseline is PlannerRFT~\citep{li2026plannerrft}, the state-of-the-art RL post-training method for diffusion-based planners, which starts from the same pretrained model and uses the same sampling setting.
The base plan spans 8\,s, the nudge policy reads its first 2\,s ($T_{\mathrm{in}}=2$\,s), and the planner replans every 0.1\,s ($\Delta t=0.1$\,s).
We also implement an open-loop finetuning from logs baseline~(\cref{sec:supp_openloop}) based on DiffusionNFT~\cite{zheng2026diffusionnft}.
Methods that train autoregressive planners with RL~\citep{zhang2025carplanner,tang2025planr1} are outside the scope of this paper, so we do not include them.
\Cref{tab:main} also lists other state-of-the-art imitation-learning methods.
All our results are presented over six runs.
Based on the score analysis in \cref{sec:supp_scores}, we call a scenario with a score below 50 a \textit{failure} and one with a score above 90 a \textit{pass}.
To build the training set, we drive the frozen planner on the training split of nuPlan in our simulator and score every drive.
The training set holds its 4,698 failures and 4,698 passes with matched scenario types, 9,396 scenarios in total.
The passes let the nudge policy learn to leave normal drives unchanged.
Both the selection of these scenarios and the training use reactive traffic.
Each run uses one NVIDIA L4 GPU.
Hyperparameters are listed in \cref{sec:supp_impl}.

\begin{figure}[t]
    \centering
    \includegraphics[width=\linewidth]{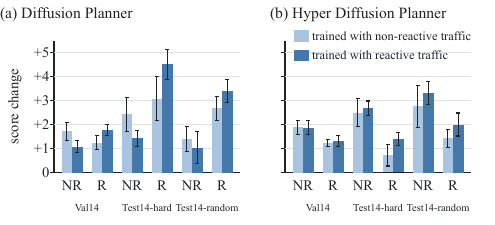}
    \vspace{-1.0\baselineskip}
    \caption{
    \textbf{\methodname on two planners and two training traffic modes, official devkit.}
    Score change over the pretrained planner, mean and standard deviation over six runs.
    }
    \label{fig:generality}
    \vspace{-0.5\baselineskip}
\end{figure}

\subsection{Main results}
\label{sec:exp_main}

\paragraph{Comparison with the state of the art.}
We compare \methodname with other methods in \cref{tab:main}.
Open-loop finetuning from logs changes the NR scores little (\rec{+0.09} / \rec{+0.65} / \rec{\textminus 0.04}) and stays below \methodname in all six settings.
In contrast, \methodname achieves clear improvement over the pretrained Diffusion Planner on both the NR (\rec{+1.07} / \rec{+1.42} / \rec{+1.04}) and the R settings (\rec{+1.78} / \rec{+4.51} / \rec{+3.39}).
Compared with PlannerRFT, \methodname achieves better or on-par performance in all six settings with about one sixth of the environment steps (6.2\,M against 40\,M).
Sub-metric comparison on Test14-hard is shown in
\cref{sec:supp_submetrics}.

\vspace{-0.2\baselineskip}
\paragraph{Qualitative results.}
\Cref{fig:qualitative} shows two scenarios (more results in \cref{sec:more_qualitative}) that the pretrained planner fails and \methodname passes.
In \crefp{fig:qualitative}{a}, a faster vehicle approaches from the rear left, and the pretrained planner collides with it at 7.0\,s.
The nudge policy slows the vehicle down, with a speed offset of up to \textminus 0.17\,m/s, and steers it to the right, with a curvature offset of up to \textminus 0.005\,m$^{-1}$, leaving the rear-left vehicle enough space to overtake.
In \crefp{fig:qualitative}{b}, the pretrained planner hits a pedestrian at 11.9\,s.
The nudge policy lowers the speed, with an offset of up to \textminus 0.36\,m/s, and steers left, with an offset of up to 0.012\,m$^{-1}$, away from the pedestrians.

\subsection{Other planners and training traffic}
\label{sec:exp_generality}

\methodname is not tied to one planner or to one type of training traffic.
In addition to the Diffusion Planner, we apply it to HDP~\citep{zheng2026hdp}, a second diffusion-based planner verified with real-vehicle tests, and we train with non-reactive or reactive traffic (\cref{fig:generality}).
Every combination improves all six settings.
The complete results are in \cref{tab:hdp}.

\subsection{How to post-train with off-policy RL}
\label{sec:exp_takeaways}
\begin{figure}[t]
    \centering
    \includegraphics[width=\linewidth]{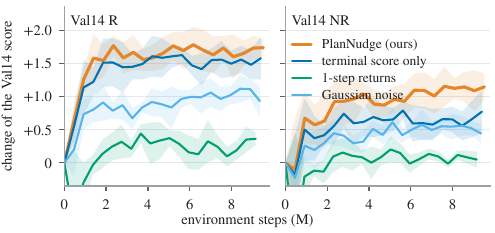}
    \vspace{-1.0\baselineskip}
    \caption{
    \textbf{Learning curves on Val14, our simulator.}
    Change over the pretrained planner against environment steps for \methodname and variants, mean and standard deviation (band) over six runs.
    }
    \label{fig:ablation_curves}
    \vspace{-1.0\baselineskip}
\end{figure}

The final scores and the sample efficiency of \methodname in \cref{sec:exp_main} show the potential of off-policy RL for post-training driving planners.
However, as far as we know, there is no established practice for off-policy RL in autonomous driving planning.
\methodname contains design choices beyond the nudge policy: what the reward scores, how far the return looks ahead under receding-horizon execution, how the exploration noise passes through the tracker, and where the maneuvers that avoid failures come from.
Some alternatives remove most of the gain, as the ablations below show.
We ablate these choices one at a time.
Results are the change over the pretrained planner over six runs in our simulator.
\Cref{fig:ablation_curves} shows the learning curves on Val14, and each paragraph below examines one variant with the quantity that shows its mechanism.
Full results are in~\cref{tab:ablation_full}.

\begin{table}[b]
\vspace{-0.5\baselineskip}  
\centering
\footnotesize
\setlength{\tabcolsep}{4pt}
\renewcommand{\arraystretch}{1.1}
\caption{
\textbf{Change over the pretrained planner on Val14 with different reward functions, our simulator.}
}
\vspace{-0.2\baselineskip}
\begin{tabular}{@{}lcccc@{}}
\toprule
& \multicolumn{2}{c}{Terminal score} & \multicolumn{2}{c}{+ Safety penalty (ours)} \\
\cmidrule(lr){2-3} \cmidrule(lr){4-5}
Change on Val14 & NR & R & NR & R \\
\midrule
Collision & \rec{+0.3} & \rec{+0.3} & \rec{+0.5} & \rec{+0.7} \\
Time-to-collision & \rec{+0.7} & \rec{+0.5} & \rec{+1.1} & \rec{+1.0} \\
Drivable & \rec{+0.4} & \rec{+0.8} & \rec{+0.5} & \rec{+0.9} \\
\midrule
Score & \rec{+0.79}\sd{0.14} & \rec{+1.62}\sd{0.13} & \rec{+1.09}\sd{0.26} & \rec{+1.78}\sd{0.27} \\
\bottomrule
\end{tabular}
\label{tab:reward}
\end{table}

\vspace{-0.2\baselineskip}
\paragraph{Shape the reward on the terms that matter.}
The closed-loop score only arrives at the end of the episode, which makes credit assignment hard.
We add a small penalty at every step when the vehicle comes close to another traffic participant or to the edge of the drivable area (\cref{sec:supp_reward}).
Adding the safety penalty improves collision, time-to-collision, and drivable scores and yields a higher overall Val14 score in both traffic modes (\cref{tab:reward}).
We recommend shaping the reward on the terms of the metric that matter most, safety here, and keeping the shaping small, so the terminal score still decides which drives count~\citep{knox2023reward}.

\begin{figure}[t]
    \centering
    \includegraphics[width=\linewidth]{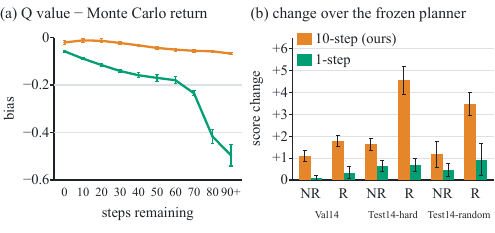}
    \vspace{-2.0\baselineskip}
    \caption{
    \textbf{Value learning with 1-step and 10-step returns, our simulator.}
    (a) Bias of the critic, its Q-value minus the Monte Carlo return of its own drives on the unseen scenarios of Val14 R, in buckets of steps remaining.
    (b) Change over the pretrained planner.
    }
    \label{fig:horizon}
    \vspace{-0.5\baselineskip}
\end{figure}

\vspace{-0.2\baselineskip}
\paragraph{Use multi-step returns.}
Even with the safety penalty, most of the reward arrives at the end of an episode of up to 150 steps.
With $1$-step targets the terminal signal reaches a state $m$ steps before the end only after $m$ rounds of bootstrapping, each adding its own approximation error~\citep{sutton2018rl}, and an $n$-step target carries it $n$ steps per update~\citep{hessel2018rainbow,fedus2020replay}.
As shown in~\cref{fig:horizon}, with $1$-step instead of 10-step returns, the critic underestimates the return of the early steps on unseen scenarios, whereas the 10-step critic stays close to the return over the whole episode with improved gain in every setting.

\begin{figure}[t]
    \centering
    \includegraphics[width=\linewidth]{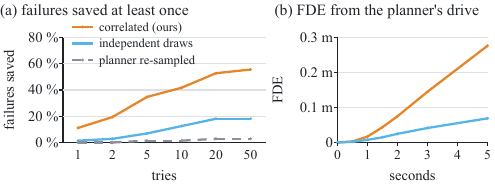}
    \vspace{-2.0\baselineskip}
    \caption{
    \textbf{Different exploration noise, our simulator.}
    The 72 Val14 NR failures of the pretrained Diffusion Planner are rerun with a zero nudge plus the noise at the training size.
    (a) Share of the failures that reach a score of at least 50 at least once.
    (b) Final displacement error (FDE) from the planner's own drive.
    }
    \label{fig:noise}
    \vspace{-0.5\baselineskip}
\end{figure}

\vspace{-0.2\baselineskip}
\paragraph{Use temporally correlated exploration noise.}
In~\cref{fig:noise}, we replace the output of the nudge policy with a zero nudge plus exploration noise and rerun the 72 Val14 NR scenarios in which the pretrained planner failed.
The correlated Ornstein--Uhlenbeck noise of \methodname saves far more failures than independent Gaussian draws of the same size, and the planner sampled again without a nudge saves almost none.
Independent draws barely move the vehicle, because the planner replans at every step and the tracker smooths the change, so the drive stays close to the planner's own drive.
Trained with independent noise, the policy learns a smaller curvature offset and its gain is lower in all settings (\cref{tab:ablation_full}).

\begin{table}[b]
\vspace{-1.0\baselineskip}  
\centering
\footnotesize
\setlength{\tabcolsep}{4pt}
\renewcommand{\arraystretch}{1.1}
\caption{
\textbf{The importance of speed events, our simulator.}
Change over the pretrained planner on Test14-hard R.
}
\vspace{-0.2\baselineskip}
\begin{tabular}{@{}lcc@{}}
\toprule
Change on Test14-hard R & Without events & + Speed events (ours) \\
\midrule
Stalls released & 9\% & 29\% \\
Progress gate & \rec{\textminus 0.1} & \rec{+0.9} \\
Progress & \rec{+0.2} & \rec{+0.3} \\
Collision & \rec{+1.2} & \rec{+1.7} \\
Time-to-collision & \rec{+1.9} & \rec{+2.1} \\
\midrule
Score & \rec{+3.25}\sd{0.34} & \rec{+4.55}\sd{0.64} \\
\bottomrule
\end{tabular}
\label{tab:events}
\end{table}

\vspace{-0.2\baselineskip}
\paragraph{Add demonstrations of the maneuvers that avoid failures.}
Two kinds of failures of the pretrained planner recur in both traffic modes.
In one, the planner keeps waiting behind a vehicle that has moved on, known as the inertia problem of behavior cloning~\citep{codevilla2019limitations}, which causes about a quarter of its failures in reactive traffic.
In the other, the planner does not slow down for a slower vehicle ahead.
A stall is released by a push, a speed increase held for seconds, and the collision is avoided by a brake held for seconds.
With a limited budget of environment steps, our agent may not explore the maneuver that these scenarios need.
Off-policy RL can learn from any demonstration: the critic values a stored transition regardless of the policy that generated the action.
We therefore execute a scripted push or brake, a speed event, in half of the collected episodes and store it like any other transition, so the critic learns the value of the maneuver and the actor adopts it where the critic values it.
On Test14-hard R, the setting with the lowest scores, the events release more stalls, the progress gate rises only with them, and the score gains more.
Scripted events are thus a targeted remedy for specific failures that exploration rarely solves.

\section{Conclusion}
\label{sec:conclusion}
We show that learning to nudge the plans of a pretrained diffusion-based planner is an effective alternative to finetuning it.
Our proposed \methodname freezes the planner and trains a nudge policy on closed-loop scores.
The policy generates a two-dimensional action in a single forward pass without iterative denoising, so the RL horizon stays short.
Both dimensions change the executed portion of the plan, which simplifies value learning under receding-horizon execution.
In all six nuPlan settings, \methodname matches or exceeds the final scores of the previous state-of-the-art RL finetuning method, with about one sixth of its environment steps and less than one fifth of its model compute.
We believe off-policy RL has great potential for post-training driving planners, and our guidelines on reward shaping, value learning, and exploration are a first step toward an established practice, and we hope that future work builds on them.


\paragraph{Limitations.}
First, a nudge is a small, bounded change of the plan.
\methodname therefore relies on a strong pretrained planner and cannot repair a plan that needs a large change.
Second, we evaluate \methodname on two diffusion-based planners that take structured scene inputs.
It is open whether nudges are as effective for an end-to-end planner that operates directly on sensor data.
Third, our closed-loop evaluation is limited to simulation.
On-road tests are needed to show whether the gains on nuPlan carry over to real vehicle dynamics and real road users.

\section{LLM Usage Statement}
We used an LLM-based coding agent, Claude Code (Anthropic), under the direction of the authors.
\textbf{Code.} The agent wrote and edited part of the training, evaluation, analysis, and plotting code.
\textbf{Experiments.} The agent launched and monitored runs, kept the experiment records, and computed the reported statistics from the logs.
\textbf{Figures and tables.} The agent drew or edited the figures and typeset the tables.
\textbf{Text.} The agent helped to check grammar and improve writing in all sections.
The authors proposed the method, verified the results, reviewed all code, figures, and text, and take full responsibility for the content.

\section*{Acknowledgments}
Zilin Wang is funded by a generous grant from Waymo.
Saeed Rahmani is partially funded by the Transport \& Mobility Institute at Delft University of Technology.
Jakob Nicolaus Foerster was partially supported by UKRI grant EP/Y028481/1 (originally selected for funding by the ERC).
Jakob Nicolaus Foerster also received support from the JPMC Research Award and the Amazon Research Award.

{
    \small
    \bibliographystyle{ieeenat_fullname}
    \bibliography{main}

@inproceedings{wagenmaker2025steering,
  author    = {Wagenmaker, Andrew and Nakamoto, Mitsuhiko and Zhang, Yunchu and Park, Seohong and Yagoub, Waleed and Nagabandi, Anusha and Gupta, Abhishek and Levine, Sergey},
  title     = {Steering Your Diffusion Policy with Latent Space Reinforcement Learning},
  booktitle = {Conference on Robot Learning},
  year      = {2025},
}

@inproceedings{ho2020ddpm,
  title     = {Denoising Diffusion Probabilistic Models},
  author    = {Ho, Jonathan and Jain, Ajay and Abbeel, Pieter},
  booktitle = {NeurIPS},
  year      = {2020}
}

@inproceedings{lipman2023flow,
  title     = {Flow Matching for Generative Modeling},
  author    = {Lipman, Yaron and Chen, Ricky T. Q. and Ben-Hamu, Heli and Nickel, Maximilian and Le, Matt},
  booktitle = {ICLR},
  year      = {2023}
}

@inproceedings{chi2023diffusionpolicy,
  title     = {Diffusion Policy: Visuomotor Policy Learning via Action Diffusion},
  author    = {Chi, Cheng and Feng, Siyuan and Du, Yilun and Xu, Zhenjia and Cousineau, Eric and Burchfiel, Benjamin and Song, Shuran},
  booktitle = {RSS},
  year      = {2023}
}

@inproceedings{zheng2025diffusionplanner,
  title     = {Diffusion-Based Planning for Autonomous Driving with Flexible Guidance},
  author    = {Zheng, Yinan and Liang, Ruiming and Zheng, Kexin and Zheng, Jinliang and Mao, Liyuan and Li, Jianxiong and Gu, Weihao and Ai, Rui and Li, Shengbo Eben and Zhan, Xianyuan and Liu, Jingjing},
  booktitle = {ICLR},
  year      = {2025}
}

@inproceedings{tan2025flowplanner,
  title     = {Flow Matching-Based Autonomous Driving Planning with Advanced Interactive Behavior Modeling},
  author    = {Tan, Tianyi and Zheng, Yinan and Liang, Ruiming and Wang, Zexu and Zheng, Kexin and Zheng, Jinliang and Li, Jianxiong and Zhan, Xianyuan and Liu, Jingjing},
  booktitle = {NeurIPS},
  year      = {2025}
}

@article{zheng2026hdp,
  title   = {Unleashing the Potential of Diffusion Models for End-to-End Autonomous Driving},
  author  = {Zheng, Yinan and Tan, Tianyi and Huang, Bin and Liu, Enguang and Liang, Ruiming and Zhang, Jianlin and Cui, Jianwei and Chen, Guang and Ma, Kun and Ye, Hangjun and Chen, Long and Zhang, Ya-Qin and Zhan, Xianyuan and Liu, Jingjing},
  journal = {arXiv preprint arXiv:2602.22801},
  year    = {2026}
}

@inproceedings{zheng2026diffusionnft,
  title     = {{DiffusionNFT}: Online Diffusion Reinforcement with Forward Process},
  author    = {Zheng, Kaiwen and Chen, Huayu and Ye, Haotian and Wang, Haoxiang and Zhang, Qinsheng and Jiang, Kai and Su, Hang and Ermon, Stefano and Zhu, Jun and Liu, Ming-Yu},
  booktitle = {ICLR},
  year      = {2026}
}

@inproceedings{karnchanachari2024nuplan,
  title     = {Towards Learning-Based Planning: The {nuPlan} Benchmark for Real-World Autonomous Driving},
  author    = {Karnchanachari, Napat and Geromichalos, Dimitris and Tan, Kok Seang and Li, Nanxiang and Eriksen, Christopher and Yaghoubi, Shakiba and Mehdipour, Noushin and Bernasconi, Gianmarco and Fong, Whye Kit and Guo, Yiluan and Caesar, Holger},
  booktitle = {ICRA},
  year      = {2024}
}

@inproceedings{dauner2023parting,
  title     = {Parting with Misconceptions about Learning-Based Vehicle Motion Planning},
  author    = {Dauner, Daniel and Hallgarten, Marcel and Geiger, Andreas and Chitta, Kashyap},
  booktitle = {CoRL},
  year      = {2023}
}

@inproceedings{ross2011dagger,
  title     = {A Reduction of Imitation Learning and Structured Prediction to No-Regret Online Learning},
  author    = {Ross, St{\'e}phane and Gordon, Geoffrey and Bagnell, Drew},
  booktitle = {AISTATS},
  year      = {2011}
}

@inproceedings{black2024ddpo,
  title     = {Training Diffusion Models with Reinforcement Learning},
  author    = {Black, Kevin and Janner, Michael and Du, Yilun and Kostrikov, Ilya and Levine, Sergey},
  booktitle = {ICLR},
  year      = {2024}
}

@inproceedings{psenka2024qsm,
  title     = {Learning a Diffusion Model Policy from Rewards via {Q}-Score Matching},
  author    = {Psenka, Michael and Escontrela, Alejandro and Abbeel, Pieter and Ma, Yi},
  booktitle = {ICML},
  year      = {2024}
}

@article{chen2025pirl,
  title   = {{$\pi_{\text{RL}}$}: Online {RL} Fine-tuning for Flow-based Vision-Language-Action Models},
  author  = {Chen, Kang and Liu, Zhihao and Zhang, Tonghe and Guo, Zhen and Xu, Si and Lin, Hao and Zang, Hongzhi and Li, Xiang and Zhang, Quanlu and Yu, Zhaofei and Fan, Guoliang and Huang, Tiejun and Wang, Yu and Yu, Chao},
  journal = {arXiv preprint arXiv:2510.25889},
  year    = {2025}
}

@inproceedings{ren2025dppo,
  title     = {Diffusion Policy Policy Optimization},
  author    = {Ren, Allen Z. and Lidard, Justin and Ankile, Lars L. and Simeonov, Anthony and Agrawal, Pulkit and Majumdar, Anirudha and Burchfiel, Benjamin and Dai, Hongkai and Simchowitz, Max},
  booktitle = {ICLR},
  year      = {2025}
}

@inproceedings{wang2023diffusionql,
  title     = {Diffusion Policies as an Expressive Policy Class for Offline Reinforcement Learning},
  author    = {Wang, Zhendong and Hunt, Jonathan J. and Zhou, Mingyuan},
  booktitle = {ICLR},
  year      = {2023}
}

@inproceedings{kang2023edp,
  title     = {Efficient Diffusion Policies for Offline Reinforcement Learning},
  author    = {Kang, Bingyi and Ma, Xiao and Du, Chao and Pang, Tianyu and Yan, Shuicheng},
  booktitle = {NeurIPS},
  year      = {2023}
}

@inproceedings{huang2025gendrive,
  title     = {{Gen-Drive}: Enhancing Diffusion Generative Driving Policies with Reward Modeling and Reinforcement Learning Fine-Tuning},
  author    = {Huang, Zhiyu and Weng, Xinshuo and Igl, Maximilian and Chen, Yuxiao and Cao, Yulong and Ivanovic, Boris and Pavone, Marco and Lv, Chen},
  booktitle = {ICRA},
  year      = {2025}
}

@article{li2025trajhf,
  title   = {Learning Personalized Driving Styles via Reinforcement Learning from Human Feedback},
  author  = {Li, Derun and Li, Changye and Wang, Yue and Ren, Jianwei and Wen, Xin and Li, Pengxiang and Xu, Leimeng and Zhan, Kun and Jia, Peng and Lang, Xianpeng and Xu, Ningyi and Zhao, Hang},
  journal = {arXiv preprint arXiv:2503.10434},
  year    = {2025}
}

@inproceedings{li2025recogdrive,
  title     = {{ReCogDrive}: A Reinforced Cognitive Framework for End-to-End Autonomous Driving},
  author    = {Li, Yongkang and Xiong, Kaixin and Guo, Xiangyu and Li, Fang and Yan, Sixu and Xu, Gangwei and Zhou, Lijun and Chen, Long and Sun, Haiyang and Wang, Bing and Ma, Kun and Chen, Guang and Ye, Hangjun and Liu, Wenyu and Wang, Xinggang},
  booktitle = {ICLR},
  year      = {2026}
}

@article{shao2024grpo,
  title   = {{DeepSeekMath}: Pushing the Limits of Mathematical Reasoning in Open Language Models},
  author  = {Shao, Zhihong and Wang, Peiyi and Zhu, Qihao and Xu, Runxin and Song, Junxiao and Bi, Xiao and Zhang, Haowei and Zhang, Mingchuan and Li, Y. K. and Wu, Y. and Guo, Daya},
  journal = {arXiv preprint arXiv:2402.03300},
  year    = {2024}
}

@inproceedings{li2026plannerrft,
  title     = {{PlannerRFT}: Reinforcing Diffusion Planners through Closed-Loop and Sample-Efficient Fine-Tuning},
  author    = {Li, Hongchen and Li, Tianyu and Yang, Jiazhi and Shang, Mingyang and Wu, Gaoqiang and Wang, Caojun and Tian, Haochen and Lin, Zengrong and Hao, Zhihui and Lang, Xianpeng and Hu, Jia and Li, Hongyang},
  booktitle = {CVPR},
  year      = {2026},
  pages     = {24929--24938}
}

@article{schulman2017ppo,
  title   = {Proximal Policy Optimization Algorithms},
  author  = {Schulman, John and Wolski, Filip and Dhariwal, Prafulla and Radford, Alec and Klimov, Oleg},
  journal = {arXiv preprint arXiv:1707.06347},
  year    = {2017}
}

@article{silver2018residual,
  title   = {Residual Policy Learning},
  author  = {Silver, Tom and Allen, Kelsey and Tenenbaum, Josh and Kaelbling, Leslie},
  journal = {arXiv preprint arXiv:1812.06298},
  year    = {2018}
}

@inproceedings{johannink2019residual,
  title     = {Residual Reinforcement Learning for Robot Control},
  author    = {Johannink, Tobias and Bahl, Shikhar and Nair, Ashvin and Luo, Jianlan and Kumar, Avinash and Loskyll, Matthias and Ojea, Juan Aparicio and Solowjow, Eugen and Levine, Sergey},
  booktitle = {ICRA},
  year      = {2019}
}

@inproceedings{ankile2024resip,
  title     = {From Imitation to Refinement -- Residual {RL} for Precise Assembly},
  author    = {Ankile, Lars and Simeonov, Anthony and Shenfeld, Idan and Torne, Marcel and Agrawal, Pulkit},
  booktitle = {ICRA},
  year      = {2025}
}

@inproceedings{yuan2025policydecorator,
  title     = {Policy Decorator: Model-Agnostic Online Refinement for Large Policy Model},
  author    = {Yuan, Xiu and Mu, Tongzhou and Tao, Stone and Fang, Yunhao and Zhang, Mengke and Su, Hao},
  booktitle = {ICLR},
  year      = {2025}
}

@inproceedings{liao2025diffusiondrive,
  title     = {{DiffusionDrive}: Truncated Diffusion Model for End-to-End Autonomous Driving},
  author    = {Liao, Bencheng and Chen, Shaoyu and Yin, Haoran and Jiang, Bo and Wang, Cheng and Yan, Sixu and Zhang, Xinbang and Li, Xiangyu and Zhang, Ying and Zhang, Qian and Wang, Xinggang},
  booktitle = {CVPR},
  year      = {2025}
}

@inproceedings{zhang2025carplanner,
  title     = {{CarPlanner}: Consistent Auto-Regressive Trajectory Planning for Large-Scale Reinforcement Learning in Autonomous Driving},
  author    = {Zhang, Dongkun and Liang, Jiaming and Guo, Ke and Lu, Sha and Wang, Qi and Xiong, Rong and Miao, Zhenwei and Wang, Yue},
  booktitle = {CVPR},
  year      = {2025}
}

@inproceedings{tang2025planr1,
  title     = {{Plan-R1}: Safe and Feasible Trajectory Planning as Language Modeling},
  author    = {Tang, Xiaolong and Kan, Meina and Shan, Shiguang and Chen, Xilin},
  booktitle = {ICLR},
  year      = {2026}
}

@inproceedings{liu2025flowgrpo,
  title     = {{Flow-GRPO}: Training Flow Matching Models via Online {RL}},
  author    = {Liu, Jie and Liu, Gongye and Liang, Jiajun and Li, Yangguang and Liu, Jiaheng and Wang, Xintao and Wan, Pengfei and Zhang, Di and Ouyang, Wanli},
  booktitle = {NeurIPS},
  year      = {2025}
}

@article{liu2025r2se,
  title   = {Reinforced Refinement with Self-Aware Expansion for End-to-End Autonomous Driving},
  author  = {Liu, Haochen and Li, Tianyu and Yang, Haohan and Chen, Li and Wang, Caojun and Guo, Ke and Tian, Haochen and Li, Hongchen and Li, Hongyang and Lv, Chen},
  journal = {IEEE TPAMI},
  year    = {2026},
  volume    = {48},
  pages     = {5774--5792}
}

@inproceedings{lillicrap2016ddpg,
  title     = {Continuous Control with Deep Reinforcement Learning},
  author    = {Lillicrap, Timothy P. and Hunt, Jonathan J. and Pritzel, Alexander and Heess, Nicolas and Erez, Tom and Tassa, Yuval and Silver, David and Wierstra, Daan},
  booktitle = {ICLR},
  year      = {2016}
}

@inproceedings{fujimoto2018td3,
  title     = {Addressing Function Approximation Error in Actor-Critic Methods},
  author    = {Fujimoto, Scott and van Hoof, Herke and Meger, David},
  booktitle = {ICML},
  year      = {2018}
}

@inproceedings{song2021sde,
  title     = {Score-Based Generative Modeling through Stochastic Differential Equations},
  author    = {Song, Yang and Sohl-Dickstein, Jascha and Kingma, Diederik P. and Kumar, Abhishek and Ermon, Stefano and Poole, Ben},
  booktitle = {ICLR},
  year      = {2021}
}

@article{lu2025dpmsolverpp,
  title     = {{DPM-Solver++}: Fast Solver for Guided Sampling of Diffusion Probabilistic Models},
  author    = {Lu, Cheng and Zhou, Yuhao and Bao, Fan and Chen, Jianfei and Li, Chongxuan and Zhu, Jun},
  journal   = {Machine Intelligence Research},
  year      = {2025},
  volume    = {22},
  pages     = {730--751}
}

@inproceedings{chen2021redq,
  title     = {Randomized Ensembled Double {Q}-Learning: Learning Fast Without a Model},
  author    = {Chen, Xinyue and Wang, Che and Zhou, Zijian and Ross, Keith},
  booktitle = {ICLR},
  year      = {2021}
}

@article{knox2023reward,
  title     = {Reward (Mis)design for Autonomous Driving},
  author    = {Knox, W. Bradley and Allievi, Alessandro and Banzhaf, Holger and Schmitt, Felix and Stone, Peter},
  journal   = {Artificial Intelligence},
  volume    = {316},
  pages     = {103829},
  year      = {2023}
}

@inproceedings{nikishin2022primacy,
  title     = {The Primacy Bias in Deep Reinforcement Learning},
  author    = {Nikishin, Evgenii and Schwarzer, Max and D'Oro, Pierluca and Bacon, Pierre-Luc and Courville, Aaron},
  booktitle = {ICML},
  year      = {2022}
}

@inproceedings{lee2025simba,
  title     = {{SimBa}: Simplicity Bias for Scaling Up Parameters in Deep Reinforcement Learning},
  author    = {Lee, Hojoon and Hwang, Dongyoon and Kim, Donghu and Kim, Hyunseung and Tai, Jun Jet and Subramanian, Kaushik and Wurman, Peter R. and Choo, Jaegul and Stone, Peter and Seno, Takuma},
  booktitle = {ICLR},
  year      = {2025}
}

@inproceedings{ankile2026resfit,
  title     = {Residual Off-Policy {RL} for Finetuning Behavior Cloning Policies},
  author    = {Ankile, Lars and Jiang, Zhenyu and Duan, Rocky and Shi, Guanya and Abbeel, Pieter and Nagabandi, Anusha},
  booktitle = {ICRA},
  pages     = {5789--5796},
  year      = {2026}
}

@inproceedings{park2025fql,
  title     = {Flow {Q}-Learning},
  author    = {Park, Seohong and Li, Qiyang and Levine, Sergey},
  booktitle = {ICML},
  year      = {2025}
}

@article{zou2025diffusiondrivev2,
  title   = {{DiffusionDriveV2}: Reinforcement Learning-Constrained Truncated Diffusion Modeling in End-to-End Autonomous Driving},
  author  = {Zou, Jialv and Chen, Shaoyu and Liao, Bencheng and Zheng, Zhiyu and Song, Yuehao and Zhang, Lefei and Zhang, Qian and Liu, Wenyu and Wang, Xinggang},
  journal = {arXiv preprint arXiv:2512.07745},
  year    = {2025}
}

@article{cheng2024pluto,
  title   = {{PLUTO}: Pushing the Limit of Imitation Learning-based Planning for Autonomous Driving},
  author  = {Cheng, Jie and Chen, Yingbing and Chen, Qifeng},
  journal = {arXiv preprint arXiv:2404.14327},
  year    = {2024}
}

@inproceedings{zhang2026dfp,
  title     = {Diffusion Forcing Planner: History-Annealed Planning with Time-Dependent Guidance for Autonomous Driving},
  author    = {Zhang, Zehan and Li, Yaoyi and Zhang, Neng and Cai, Jia},
  booktitle = {CVPR},
  pages     = {39796--39805},
  year      = {2026}
}

@inproceedings{codevilla2019limitations,
  title = {Exploring the Limitations of Behavior Cloning for Autonomous Driving},
  author = {Codevilla, Felipe and Santana, Eder and L{\'o}pez, Antonio M. and Gaidon, Adrien},
  booktitle = {ICCV},
  year = {2019}
}

@inproceedings{ball2023efficient,
  title={Efficient online reinforcement learning with offline data},
  author={Ball, Philip J and Smith, Laura and Kostrikov, Ilya and Levine, Sergey},
  booktitle={International Conference on Machine Learning},
  pages={1577--1594},
  year={2023},
  organization={PMLR}
}

@article{vecerik2017leveraging,
  title={Leveraging demonstrations for deep reinforcement learning on robotics problems with sparse rewards},
  author={Vecerik, Mel and Hester, Todd and Scholz, Jonathan and Wang, Fumin and Pietquin, Olivier and Piot, Bilal and Heess, Nicolas and Roth{\"o}rl, Thomas and Lampe, Thomas and Riedmiller, Martin},
  journal={arXiv preprint arXiv:1707.08817},
  year={2017}
}

@inproceedings{
wang2026learning,
title={Learning to Drive in New Cities Without Human Demonstrations},
author={Wang, Zilin and Rahmani, Saeed and Cornelisse, Daphne and Sarkar, Bidipta and Goldie, Alexander David and Foerster, Jakob Nicolaus and Whiteson, Shimon},
booktitle={The Fortieth Annual Conference on Neural Information Processing Systems},
year={2026},
url={https://openreview.net/forum?id=VoQnv8zzSL}
}

@inproceedings{sun2024causalagents,
  title={{CausalAgents}: A robustness benchmark for motion forecasting},
  author={Sun, Liting and Roelofs, Rebecca and Caine, Ben and Refaat, Khaled S. and Sapp, Ben and Ettinger, Scott and Chai, Wei},
  booktitle={2024 IEEE International Conference on Robotics and Automation (ICRA)},
  pages={6820--6827},
  year={2024},
  organization={IEEE}
}

@inproceedings{
cusumano-towner2025robust,
title={Robust Autonomy Emerges from Self-Play},
author={Marco Francis Cusumano-Towner and David Hafner and Alexander Hertzberg and Brody Huval and Aleksei Petrenko and Eugene Vinitsky and Erik Wijmans and Taylor W. Killian and Stuart Bowers and Ozan Sener and Philipp Kraehenbuehl and Vladlen Koltun},
booktitle={Forty-second International Conference on Machine Learning},
year={2025},
url={https://openreview.net/forum?id=yOXoJpG6Qy}
}

@article{seyam2026model,
  title={Model-Aware Generation of Safety-Critical Scenes for Robust End-to-End Driving via Distilled Planner Surrogates},
  author={Seyam, Khaled M. and Wiederer, Julian and Fathalah, Samar and Braun, Markus and Yang, Bin},
  journal={arXiv preprint arXiv:2610.35448},
  year={2026}
}

@inproceedings{
ma2026what,
title={What Makes Value Learning Efficient in Residual Reinforcement Learning?},
author={Guozheng Ma and Lu Li and Haoyu Wang and Zixuan Liu and Pierre-Luc Bacon and Dacheng Tao},
booktitle={Forty-third International Conference on Machine Learning},
year={2026},
url={https://openreview.net/forum?id=tSMZHXtJwZ}
}

@inproceedings{song2021ddim,
  title = {Denoising Diffusion Implicit Models},
  author = {Song, Jiaming and Meng, Chenlin and Ermon, Stefano},
  booktitle = {ICLR},
  year = {2021}
}

@inproceedings{cui2019multimodal,
  title={Multimodal trajectory predictions for autonomous driving using deep convolutional networks},
  author={Cui, Henggang and Radosavljevic, Vladan and Chou, Fang-Chieh and Lin, Tsung-Han and Nguyen, Thi and Huang, Tzu-Kuo and Schneider, Jeff and Djuric, Nemanja},
  booktitle={2019 International Conference on Robotics and Automation (ICRA)},
  pages={2090--2096},
  year={2019},
  organization={IEEE}
}

@inproceedings{rhinehart2019precog,
  title={{PRECOG}: Prediction conditioned on goals in visual multi-agent settings},
  author={Rhinehart, Nicholas and McAllister, Rowan and Kitani, Kris and Levine, Sergey},
  booktitle={2019 IEEE/CVF International Conference on Computer Vision (ICCV)},
  pages={2821--2830},
  year={2019},
  organization={IEEE}
}

@INPROCEEDINGS{cheng2024rethinking,
  author={Cheng, Jie and Chen, Yingbing and Mei, Xiaodong and Yang, Bowen and Li, Bo and Liu, Ming},
  booktitle={2024 IEEE International Conference on Robotics and Automation (ICRA)}, 
  title={Rethinking Imitation-based Planners for Autonomous Driving}, 
  year={2024},
  pages={14123-14130},
  doi={10.1109/ICRA57147.2024.10611364}}

@article{Treiber_2000,
   title={Congested traffic states in empirical observations and microscopic simulations},
   volume={62},
   ISSN={1095-3787},
   url={http://dx.doi.org/10.1103/PhysRevE.62.1805},
   DOI={10.1103/physreve.62.1805},
   number={2},
   journal={Physical Review E},
   publisher={American Physical Society (APS)},
   author={Treiber, Martin and Hennecke, Ansgar and Helbing, Dirk},
   year={2000},
   month=Aug, pages={1805–1824} }

@book{sutton2018rl,
  title     = {Reinforcement Learning: An Introduction},
  author    = {Sutton, Richard S. and Barto, Andrew G.},
  publisher = {MIT Press},
  edition   = {2nd},
  year      = {2018}
}

@inproceedings{hessel2018rainbow,
  title     = {Rainbow: Combining Improvements in Deep Reinforcement Learning},
  author    = {Hessel, Matteo and Modayil, Joseph and van Hasselt, Hado and Schaul, Tom and Ostrovski, Georg and Dabney, Will and Horgan, Dan and Piot, Bilal and Azar, Mohammad and Silver, David},
  booktitle = {AAAI},
  year      = {2018}
}

@inproceedings{fedus2020replay,
  title     = {Revisiting Fundamentals of Experience Replay},
  author    = {Fedus, William and Ramachandran, Prajit and Agarwal, Rishabh and Bengio, Yoshua and Larochelle, Hugo and Rowland, Mark and Dabney, Will},
  booktitle = {ICML},
  year      = {2020}
}

@inproceedings{dehaan2019causal,
  title     = {Causal Confusion in Imitation Learning},
  author    = {de Haan, Pim and Jayaraman, Dinesh and Levine, Sergey},
  booktitle = {Advances in Neural Information Processing Systems},
  volume    = {32},
  year      = {2019}
}
}

\clearpage
\newpage
\twocolumn[{
  \centering
  {\Large \textbf{\thetitle}\\ \vspace{0.5em}Supplementary Material \\ \vspace{1.0em}}
  \includegraphics[width=\textwidth]{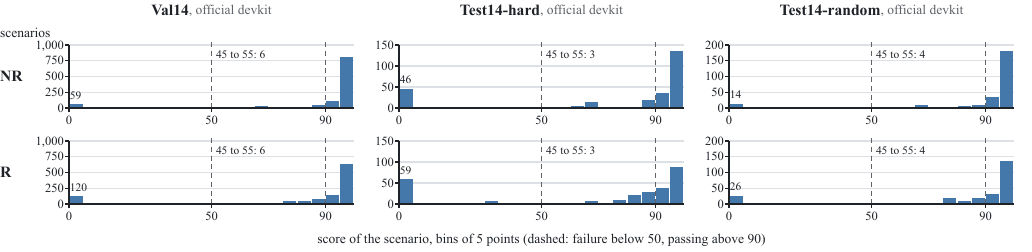}
  \captionof{figure}{
    \textbf{Scores of the frozen Diffusion Planner across the six evaluation settings.}
    Columns correspond to the three scenario sets, evaluated using the official devkit with one run of the frozen planner.
    Rows correspond to non-reactive (NR) and reactive (R) traffic.
    Scores are grouped into bins of 5 points.
    The number above the first bar counts scenarios scoring exactly 0, and the label at the top counts scenarios scoring between 45 and 55.
    Dashed lines mark the failure threshold at 50 and the pass threshold at 90.
  }
  \label{fig:supp_scores}
  \vspace{1.0em}
}]
\setcounter{section}{0}
\renewcommand{\thesection}{\Alph{section}}
\renewcommand{\theHsection}{supp.\Alph{section}}

\begin{table*}[t]
\centering
\footnotesize
\setlength{\tabcolsep}{2.7pt}
\caption{Hyperparameters of \methodname.}
\begin{tabular}{@{}lr@{\hspace{0.6em}}lr@{\hspace{0.6em}}lr@{}}
\toprule
Parameter & Value & Parameter & Value & Parameter & Value \\
\midrule
\multicolumn{2}{@{}l}{\textit{Nudge policy}} & \multicolumn{2}{@{}l}{\textit{RL}} & \multicolumn{2}{@{}l}{\textit{Data collection}} \\
Input: pooled encoder tokens & 192 & Critics $M$ & 10 & Environments & 32 \\
Input: ego state & 10 & Target subset $|\mathcal S|$ & 2 & Episodes per iteration & 32 \\
Input: first 2\,s of the plan & $20\times4$ & Discount $\gamma$ & 0.99 & Episode length & at most 150 steps \\
Speed bound $v_{\max}$ & 1.0\,m/s & $n$-step & 10 & Control rate & 10\,Hz \\
Curvature bound $\kappa_{\max}$ & 0.05\,m$^{-1}$ & Batch size & 2,048 & Warm-up transitions & 20,000 \\
Lateral acceleration bound $a^{\mathrm{lat}}_{\max}$ & 1.5\,m/s$^2$ & Update-to-data ratio & 0.0125 & Iterations & 2,240 \\
Speed ramp $t_{\mathrm{ramp}}$ & 1.1\,s & Learning rate, actor & $1\times10^{-4}$ & Replay size & 11\,M transitions \\
Curvature window $H_\kappa$ & 15 steps & Learning rate, critic & $3\times10^{-4}$ & Exploration noise & Ornstein--Uhlenbeck \\
Network & SimBa residual MLP & Policy delay & 2 & Exploration noise size & 0.1 of the bound \\
Residual blocks & 2 & Polyak $\tau$ & 0.005 & Exploration noise time constant & 1.5\,s \\
Actor width & 256 & Target noise & 0.2 & & \\
Critic width & 512 & Target noise clip & 0.5 & & \\
\cmidrule(r){1-2} \cmidrule(r){3-4} \cmidrule(r){5-6}
\multicolumn{2}{@{}l}{\textit{Base planners}} & \multicolumn{2}{@{}l}{\textit{Action penalty}} & \multicolumn{2}{@{}l}{\textit{Critic resets}} \\
Sampler, HDP & DPM-Solver++, order 2 & Speed term & $0.05\,(a^{v})^2$ & Critic-only warm-up & 1,250 updates \\
Sampler, Diffusion Planner & DDIM & Curvature term & $0.125\,(a^{\kappa})^2$ & Reset period & 448 iterations \\
Solver steps & 5 & & & Updates after a reset & 25,000 \\
Initial noise scale & 0.1 & & & & \\
\bottomrule
\end{tabular}
\label{tab:hyper}
\end{table*}

\section{Failure analysis}
\label{sec:analysis}

This section presents the failure analysis that motivates \methodname
(\cref{sec:intro}).
Unless otherwise stated, results are obtained using the frozen Diffusion
Planner~\citep{zheng2025diffusionplanner} on the 1,118 Val14 NR scenarios
in our simulator.
We classify a scenario as a failure if its closed-loop score is below 50.

\paragraph{Few scenarios lie near the failure threshold.}
\label{sec:supp_scores}
We examine the pretrained planner's score distribution across all six evaluation settings.
\Cref{fig:supp_scores} shows that few scenarios have scores near the threshold of 50.
Most failures receive a score of 0 because at least one of the score's four gating terms is zero.
At this threshold, the planner fails in 72 Val14 NR scenarios in our simulator: 38 due to at-fault collisions, 16 due to drivable-area violations, and the remaining 18 due to low scores for other reasons.

\begin{figure}[b]
    \centering
    \includegraphics[width=1.0\linewidth]{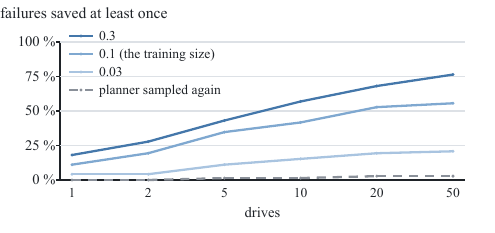}
    \caption{
    \textbf{Temporally correlated random nudges can avoid closed-loop failures.}
    For each condition, we perform 50 closed-loop rollouts in our simulator for each of the 72 Val14 NR scenarios in which the frozen Diffusion Planner fails.
    For a rollout budget of $k$, the curves show the fraction of these scenarios in which at least one of the $k$ rollouts achieves a score of at least 50.
    Solid lines correspond to temporally correlated random nudges to the planned speed and curvature at three noise scales, expressed relative to the corresponding action bounds.
    The dashed line resamples the planner with new diffusion noise, without applying a nudge.
    }
    \label{fig:saves}
\end{figure}

\paragraph{Random nudges reveal opportunities to avoid failures.}
We perturb each plan using the action space of \methodname (\cref{sec:method_system}).
The speed offset is bounded by 1\,m/s, and the curvature offset at the beginning of the plan is bounded by 0.05\,m$^{-1}$, subject to the constraints of the plan transform (\cref{sec:supp_transform}).
The nudges are random and temporally correlated across planning steps (\cref{sec:method_training}).
For each noise scale, we rerun each of the 72 failure scenarios 50 times, using new random nudges in each rollout.
At a noise scale of 0.1 relative to the action bounds, 53\% of the initially failed scenarios achieve a score of at least 50 in at least one of 20 rollouts.
This fraction increases with the noise scale (\cref{fig:saves}).
By comparison, resampling the planner with new diffusion noise and no nudge avoids failure in only 2 of the 72 scenarios, even after 50 rollouts per scenario.
Under the sampling configuration used here, resampling alone therefore rarely resolves these failures.

\section{Open-loop scores miss closed-loop failures}
\label{sec:plan_drive}

The planner makes a new plan at every planning step and the tracker follows the newest one, so the driven trajectory lies on none of the plans.
A plan can therefore be safe in open loop while the drive fails in closed loop.
We test this on the 54 failures of \cref{sec:analysis} by a collision or a road departure.
In the open-loop test, the vehicle follows a plan as it is, without replanning and without a tracker, against the logged motion of the other traffic participants, over the whole plan of 8\,s or up to the end of the scenario.
In 15 of the 54 failures, the plan made 0.5\,s before the failure has no collision and no road departure in this test, although the closed-loop drive fails 0.5\,s later.
The count is the same for a window of 4\,s, the horizon of the reward of PlannerRFT~\citep{li2026plannerrft} and of our open-loop baseline (\cref{sec:supp_openloop}).
An open-loop score misses these failures, while a score of the driven trajectory sees all of them.

\section{Reward}
\label{sec:supp_reward}

The episode reward combines a terminal score with a safety penalty at each step.
At the end of the episode, the environment computes the closed-loop score $S \in [0, 100]$ of the driven trajectory using the nuPlan devkit's metric definitions.
The terminal score contribution is $S/100$.
An at-fault collision or a departure more than 0.3\,m beyond the drivable area terminates the episode with a terminal score contribution of zero.
At every step $t$, the environment also adds the safety penalty
\begin{equation}
    r^{\mathrm{m}}_t = -0.02 \big( p^{\mathrm{col}}_t + p^{\mathrm{off}}_t \big).
    \label{eq:margin}
\end{equation}
Both penalty terms use $\mathrm{ramp}_{x_0}(x) = \mathrm{clip}\big((x_0 - x)/x_0, 0, 1\big)$, which is zero when $x\geq x_0$ and one when $x=0$.
The collision term is zero when the ego speed is below 0.05\,m/s.
Otherwise, it is the larger of $\mathrm{ramp}_{2\,\mathrm{s}}(\mathrm{TTC})$ and $\mathrm{ramp}_{0.5\,\mathrm{m}}(\mathrm{gap})$.
Here, $\mathrm{TTC}$ is the time to collision with a traffic participant ahead, assuming constant velocities, and $\mathrm{gap}$ is the smallest distance to a traffic participant beside the ego vehicle.
Traffic participants with which the ego vehicle has already made contact are excluded, as in the devkit's at-fault collision rule.

The drivable-area term is $\mathrm{ramp}_{1\,\mathrm{s}}(\mathrm{TTO})$.
Here, $\mathrm{TTO}$ is the time for the vehicle corner closest to the drivable-area boundary to exceed the 0.3\,m tolerance, assuming its current speed toward the boundary remains constant.
We set $p_t^{\mathrm{off}}=0$ when this outward speed is below 0.05\,m/s.
We choose the thresholds so that driving by the logged expert and by the planner in passing scenarios incurs almost no penalty.
Evaluation uses the unshaped score $S$.

\section{Implementation details}
\label{sec:supp_impl}

\Cref{tab:hyper} lists the hyperparameters.

\subsection{Plan transform}
\label{sec:supp_transform}

\begin{table*}[t]
\centering
\small
\setlength{\tabcolsep}{3pt}
\renewcommand{\arraystretch}{1.15}
\caption{
\textbf{\methodname on two planners and two training traffic modes, official devkit.}
\methodname rows: the mean over six training runs with its standard deviation after it, each run evaluated as its own policy and its score rounded to two decimals before averaging. In brackets, the change of the mean over the pretrained planner, whose Diffusion Planner row is the mean over six devkit runs with different sampling-noise seeds.
The last column counts the environment steps of RL training up to the reported checkpoint.
}
\begin{tabular}{@{}lccccccc@{}}
\toprule
& \multicolumn{2}{c}{Val14} & \multicolumn{2}{c}{Test14-hard} & \multicolumn{2}{c}{Test14-random} & RL \\
\cmidrule(lr){2-3} \cmidrule(lr){4-5} \cmidrule(lr){6-7}
Method & NR & R & NR & R & NR & R & steps \\
\midrule
HDP~\citep{zheng2026hdp} (our evaluation) & \rec{89.74} & \rec{82.18} & \rec{75.91} & \rec{69.06} & \rec{88.02} & \rec{81.94} & -- \\
\methodname, trained with non-reactive traffic & \cellms{\rec{91.62}}{\rec{0.28}}{\rec{+1.88}} & \cellms{\rec{83.41}}{\rec{0.15}}{\rec{+1.23}} & \cellms{\rec{78.41}}{\rec{0.58}}{\rec{+2.50}} & \cellms{\rec{69.77}}{\rec{0.44}}{\rec{+0.71}} & \cellms{\rec{90.78}}{\rec{0.88}}{\rec{+2.76}} & \cellms{\rec{83.36}}{\rec{0.36}}{\rec{+1.42}} & \rec{4.6\,M} \\
\methodname, trained with reactive traffic & \cellms{\rec{91.61}}{\rec{0.28}}{\rec{+1.87}} & \cellms{\rec{83.49}}{\rec{0.24}}{\rec{+1.31}} & \cellms{\rec{78.59}}{\rec{0.30}}{\rec{+2.68}} & \cellms{\rec{70.46}}{\rec{0.27}}{\rec{+1.40}} & \cellms{\rec{91.35}}{\rec{0.48}}{\rec{+3.33}} & \cellms{\rec{83.94}}{\rec{0.48}}{\rec{+2.00}} & \rec{8.9\,M} \\
\midrule
Diffusion Planner, DDIM (our evaluation) & \rec{90.00} & \rec{83.22} & \rec{75.74} & \rec{68.93} & \rec{89.99} & \rec{82.70} & -- \\
\methodname, trained with non-reactive traffic & \cellms{\rec{91.72}}{\rec{0.37}}{\rec{+1.72}} & \cellms{\rec{84.47}}{\rec{0.28}}{\rec{+1.25}} & \cellms{\rec{78.16}}{\rec{0.71}}{\rec{+2.42}} & \cellms{\rec{72.00}}{\rec{0.92}}{\rec{+3.07}} & \cellms{\rec{91.38}}{\rec{0.53}}{\rec{+1.39}} & \cellms{\rec{85.37}}{\rec{0.51}}{\rec{+2.67}} & \rec{7.7\,M} \\
\methodname, trained with reactive traffic & \cellms{\rec{91.07}}{\rec{0.25}}{\rec{+1.07}} & \cellms{\rec{85.00}}{\rec{0.23}}{\rec{+1.78}} & \cellms{\rec{77.16}}{\rec{0.32}}{\rec{+1.42}} & \cellms{\rec{73.44}}{\rec{0.63}}{\rec{+4.51}} & \cellms{\rec{91.03}}{\rec{0.65}}{\rec{+1.04}} & \cellms{\rec{86.09}}{\rec{0.47}}{\rec{+3.39}} & \rec{6.2\,M} \\
\bottomrule
\end{tabular}
\label{tab:hdp}
\end{table*}

We implement the plan transform $T$ shown in \cref{fig:method} in three steps.
The transform applies a normalized nudge $a=(a^{v},a^\kappa)\in[-1,1]^2$ to the base plan.
We omit the planning-step index and use a bar for base-plan quantities and a tilde for adjusted quantities.
For $h=1,\ldots,H$, base waypoint $h$ has position $\bar{\mathbf p}^{h}=(\bar x^{h},\bar y^{h})$ and heading $\bar\theta^{h}$ in the current ego frame, with $\bar{\mathbf p}^{0}=\mathbf 0$.
Consecutive waypoints are separated by $\delta=0.1$\,s.
We define $\bar{\mathbf d}^h=\bar{\mathbf p}^{h}-\bar{\mathbf p}^{h-1}$ as the displacement between consecutive waypoints, which we call step $h$, and $\bar{\mathbf e}^h=(\cos\bar\theta^h,\sin\bar\theta^h)$ as the unit heading direction at waypoint $h$.

\paragraph{Step 1 speed.}
The speed component $a^{v}$ sets a speed offset that grows linearly to $a^{v}v_{\max}$ within the ramp time $t_{\mathrm{ramp}}=1.1$\,s and then stays constant,
\begin{equation}
    \Delta v^h=a^{v}\,v_{\max}\min\!\Big(\frac{h\delta}{t_{\mathrm{ramp}}},1\Big),
    \label{eq:nudge_speed}
\end{equation}
with the speed bound $v_{\max}=1$\,m/s.
Let $\bar s^h=\bar{\mathbf e}^{h\top}\bar{\mathbf d}^h$ be the forward length of base step $h$, the distance that the vehicle moves along its heading in this step.
Without a limit, the speed offset would change this length by $\Delta v^h\delta$.
A brake may reduce the forward length to zero and not below, so the applied change $u^h$ and the nudged step are
\begin{equation}
    \begin{aligned}
        u^h&=\max\!\big(\Delta v^h\delta,\,-\max(\bar s^h,0)\big),\\
        \tilde{\mathbf d}^h&=\bar{\mathbf d}^h+u^h\,\bar{\mathbf e}^h.
    \end{aligned}
    \label{eq:nudge_acc}
\end{equation}
For example, with $\Delta v^h\delta=-0.1$\,m, a step of 0.5\,m becomes 0.4\,m, and a step of 0.03\,m becomes 0\,m.

\paragraph{Step 2 curvature.}
The curvature component $a^\kappa$ sets a curvature offset
\begin{equation}
    \Delta\kappa=a^\kappa\min\!\left(
    \kappa_{\max},\frac{a^{\mathrm{lat}}_{\max}}{\max(\bar v,v_{\min})^2}
    \right),
    \label{eq:curvature}
\end{equation}
where $\kappa_{\max}=0.05$\,m$^{-1}$ is the curvature bound and $\bar v=\frac{1}{10\delta}\sum_{j=1}^{10}\|\bar{\mathbf d}^j\|$ is the mean speed of the base plan over its first second, the length of its first ten steps divided by their duration.
The second term keeps the added lateral acceleration below $a^{\mathrm{lat}}_{\max}=1.5$\,m/s$^2$ at high speed, and $v_{\min}=10^{-3}$\,m/s avoids a division by zero.
A curvature offset turns the path by $\Delta\kappa$ per meter of traveled distance.
The turn is applied over the first $H_\kappa=15$ steps (1.5\,s), so the heading offset at waypoint $h$ is
\begin{equation}
    \varphi^h=\Delta\kappa
    \sum_{j=1}^{\min(h,H_\kappa)}\big\|\tilde{\mathbf d}^j\big\|,
    \label{eq:nudge_heading}
\end{equation}
and it stays constant after step $H_\kappa$.

\paragraph{Step 3 nudged plan.}
We construct the nudged plan by rotating each speed-adjusted displacement by its heading offset and adding it to the preceding nudged waypoint:
\begin{equation}
    \tilde{\mathbf p}^h=\tilde{\mathbf p}^{h-1}+R(\varphi^h)\,\tilde{\mathbf d}^h,
    \qquad
    \tilde\theta^h=\bar\theta^h+\varphi^h,
    \label{eq:nudge_curv}
\end{equation}
with $\tilde{\mathbf p}^{0}=\mathbf 0$ and $R(\varphi)$ the rotation by the angle $\varphi$.
Because the heading offset is constant after step $H_\kappa$, the rest of the plan keeps its shape and is only rotated.
For a zero nudge, $u^h=0$ and $\varphi^h=0$, so the transform returns the base plan exactly.

\begin{table*}[t]
\centering
\small
\setlength{\tabcolsep}{3pt}
\renewcommand{\arraystretch}{1.0}
\caption{
\textbf{The six settings of every ablation row, our simulator.}
Diffusion Planner with the DDIM solver.
Each cell: change over the pretrained planner, mean $\pm$ SD over runs, and in brackets the failures fixed (new score at least 50) / the other scenarios broken (new score below 50) per run.
Six runs per row.
}
\begin{tabular}{@{}lcccccc@{}}
\toprule
& \multicolumn{2}{c}{Val14} & \multicolumn{2}{c}{Test14-hard} & \multicolumn{2}{c}{Test14-random} \\
\cmidrule(lr){2-3} \cmidrule(lr){4-5} \cmidrule(lr){6-7}
Variant & NR & R & NR & R & NR & R \\
\midrule
\methodname (ours) & \cellsd{\rec{+1.09}}{\rec{0.26}}{\rec{19.8 / 5.5}} & \cellsd{\rec{+1.78}}{\rec{0.27}}{\rec{28.8 / 6.5}} & \cellsd{\rec{+1.63}}{\rec{0.27}}{\rec{7.8 / 3.5}} & \cellsd{\rec{+4.55}}{\rec{0.64}}{\rec{16.2 / 1.3}} & \cellsd{\rec{+1.17}}{\rec{0.58}}{\rec{5.0 / 3.3}} & \cellsd{\rec{+3.45}}{\rec{0.52}}{\rec{9.8 / 0.3}} \\
\midrule
\multicolumn{7}{@{}l}{\textit{(a) Reward}} \\
Terminal score only & \cellsd{\rec{+0.79}}{\rec{0.14}}{\rec{13.5 / 4.0}} & \cellsd{\rec{+1.62}}{\rec{0.13}}{\rec{24.0 / 4.3}} & \cellsd{\rec{+1.64}}{\rec{0.42}}{\rec{5.8 / 1.8}} & \cellsd{\rec{+3.57}}{\rec{0.81}}{\rec{12.2 / 1.3}} & \cellsd{\rec{+1.14}}{\rec{0.72}}{\rec{4.0 / 1.7}} & \cellsd{\rec{+3.38}}{\rec{0.28}}{\rec{10.3 / 0.5}} \\
\midrule
\multicolumn{7}{@{}l}{\textit{(b) Value learning}} \\
1-step returns & \cellsd{\rec{+0.08}}{\rec{0.12}}{\rec{4.0 / 3.0}} & \cellsd{\rec{+0.34}}{\rec{0.28}}{\rec{6.7 / 2.2}} & \cellsd{\rec{+0.65}}{\rec{0.25}}{\rec{1.8 / 0.2}} & \cellsd{\rec{+0.70}}{\rec{0.30}}{\rec{2.5 / 0.7}} & \cellsd{\rec{+0.44}}{\rec{0.30}}{\rec{1.5 / 0.7}} & \cellsd{\rec{+0.93}}{\rec{0.73}}{\rec{2.5 / 0.0}} \\
\midrule
\multicolumn{7}{@{}l}{\textit{(c) Exploration}} \\
Gaussian noise & \cellsd{\rec{+0.52}}{\rec{0.11}}{\rec{10.7 / 4.8}} & \cellsd{\rec{+1.11}}{\rec{0.15}}{\rec{17.2 / 3.5}} & \cellsd{\rec{+1.45}}{\rec{0.48}}{\rec{5.5 / 2.0}} & \cellsd{\rec{+2.78}}{\rec{0.75}}{\rec{9.7 / 0.7}} & \cellsd{\rec{+0.42}}{\rec{0.55}}{\rec{2.3 / 1.8}} & \cellsd{\rec{+1.96}}{\rec{0.69}}{\rec{5.3 / 0.3}} \\
No speed events & \cellsd{\rec{+1.09}}{\rec{0.14}}{\rec{20.3 / 7.3}} & \cellsd{\rec{+2.01}}{\rec{0.19}}{\rec{29.7 / 5.3}} & \cellsd{\rec{+1.22}}{\rec{0.38}}{\rec{7.2 / 4.3}} & \cellsd{\rec{+3.25}}{\rec{0.34}}{\rec{11.7 / 1.5}} & \cellsd{\rec{+1.02}}{\rec{0.52}}{\rec{4.5 / 3.5}} & \cellsd{\rec{+3.31}}{\rec{0.29}}{\rec{9.8 / 0.7}} \\
\bottomrule
\end{tabular}
\label{tab:ablation_full}
\vspace{6pt}
\end{table*}

\subsection{Actor-critic updates}
\label{sec:supp_updates}

Episodes have a fixed maximum length, and the terminal score is received only at the end.
To account for the dependence of value on the remaining time, we use an asymmetric actor-critic architecture: the critics receive the remaining episode fraction $\rho_t=(150-t)/150$, stored in each transition, as a privileged input.
The actor does not receive this input.
The critics learn from $n$-step returns with $n=10$ (\cref{tab:hyper}).
For a sampled segment of length $\ell\leq n$, truncated at the episode boundary, the critic target is
\begin{equation}
    \begin{aligned}
        y_t={}&\sum_{i=0}^{\ell-1}\gamma^i r_{t+i} \\
        &+\gamma^\ell b_t\min_{j\in\mathcal S}
        Q_{\bar\psi_j}(z_{t+\ell},\rho_{t+\ell},a'),
    \end{aligned}
    \label{eq:target}
\end{equation}
where $b_t=0$ if the segment reaches the episode end and $b_t=1$ otherwise.
Both safety termination and the 15\,s scenario boundary are terminal, with no bootstrapping.
The set $\mathcal S$ contains two target critics sampled without replacement at each update.
The target action is $a'=\mathrm{clip}(\mu_{\bar\phi}(z_{t+\ell})+\epsilon,-1,1)$, with clipped Gaussian noise $\epsilon$.
Each critic minimizes the squared error to $y_t$.
The actor follows \cref{eq:actor_loss}, whose penalty $\lambda\|Wa\|_2^2$ is $0.05\,(a^{v})^2+0.125\,(a^{\kappa})^2$ on the normalized action components.

After a critic-only warm-up, we update the actor once every two critic updates.
Target networks are updated using Polyak averaging.
We periodically reinitialize the critics and their target networks while retaining the actor and replay buffer.
Before resuming data collection, we fit the new critics to stored transitions~\citep{nikishin2022primacy}.
The warm-up, reset schedule, and other settings are listed in \cref{tab:hyper}.

\subsection{Speed events}
\label{sec:supp_events}

Speed events are enabled in each data-collection episode with probability 0.5.
In these episodes, while a push or a brake is active, its value replaces the exploration noise added to the actor's speed component.
The curvature component always retains its exploration noise.
A push starts when the vehicle has stood for 1\,s, its base plan covers less than 0.2\,m in the next second, and the IDM would accelerate the vehicle.
Specifically, we apply the IDM to the ego vehicle with a standstill gap of 1\,m and require an acceleration above 0.3\,m/s$^2$.
The push adds up to half of the speed bound and lasts while the IDM still gives this acceleration.
A brake is held for a duration drawn uniformly from 2 to 5\,s and lowers the planned speed by a quarter of the current speed, at most by the speed bound.
The first hold starts with the episode.
When a hold ends, the next hold starts at once with probability 0.5 and after a pause drawn from the same range otherwise.
No brake acts during a push.
The replay buffer stores the executed action of every step.

\begin{table*}[t]
\centering
\small
\setlength{\tabcolsep}{1.3pt}
\renewcommand{\arraystretch}{1.15}
\caption{
\textbf{Sub-metrics on Test14-hard, official devkit.}
Bold denotes the highest result of each column, excluding the log replay.
Red: a change below zero. Blue: a change above 1 point.
}
\begin{tabular}{@{}lcccccccc|c@{}}
\toprule
Method & \textbf{R score} & Collision & TTC & Drivable & Progress gate & Progress & Speed & Comfort & \textbf{NR score} \\
\midrule
\textcolor{gray}{Log replay} & \textcolor{gray}{68.80} & \textcolor{gray}{77.02} & \textcolor{gray}{69.85} & \textcolor{gray}{95.96} & \textcolor{gray}{100.00} & \textcolor{gray}{98.48} & \textcolor{gray}{94.12} & \textcolor{gray}{99.26} & \textcolor{gray}{85.96} \\
\midrule
Pretrained planner & \rec{68.93} & \rec{88.02} & \rec{80.57} & \rec{95.28} & \rec{91.24} & \rec{75.59} & \rec{97.55} & \rec{87.44} & \rec{75.74} \\
Open-loop finetuning from logs & \cellms{\rec{70.42}}{\rec{0.20}}{\textcolor{improve}{\rec{+1.49}}} & \cellms{\rec{86.76}}{\rec{0.17}}{\textcolor{degrade}{\rec{\textminus 1.26}}} & \cellms{\rec{79.29}}{\rec{0.30}}{\textcolor{degrade}{\rec{\textminus 1.28}}} & \cellms{\rec{95.41}}{\rec{0.31}}{\rec{+0.13}} & \cellms{\textbf{\rec{94.24}}}{\rec{0.19}}{\textcolor{improve}{\rec{+3.00}}} & \cellms{\textbf{\rec{78.67}}}{\rec{0.42}}{\textcolor{improve}{\rec{+3.08}}} & \cellms{\rec{97.53}}{\rec{0.13}}{\textcolor{degrade}{\rec{\textminus 0.02}}} & \cellms{\textbf{\rec{88.66}}}{\rec{0.28}}{\textcolor{improve}{\rec{+1.22}}} & \cellms{\rec{76.39}}{\rec{0.89}}{\rec{+0.65}} \\
\methodname (ours) & \cellms{\textbf{\rec{73.44}}}{\rec{0.63}}{\textcolor{improve}{\rec{+4.51}}} & \cellms{\textbf{\rec{90.04}}}{\rec{0.47}}{\textcolor{improve}{\rec{+2.02}}} & \cellms{\textbf{\rec{83.58}}}{\rec{0.68}}{\textcolor{improve}{\rec{+3.01}}} & \cellms{\textbf{\rec{97.00}}}{\rec{0.15}}{\textcolor{improve}{\rec{+1.72}}} & \cellms{\rec{93.32}}{\rec{0.43}}{\textcolor{improve}{\rec{+2.08}}} & \cellms{\rec{76.18}}{\rec{0.31}}{\rec{+0.59}} & \cellms{\textbf{\rec{97.58}}}{\rec{0.05}}{\rec{+0.03}} & \cellms{\rec{87.69}}{\rec{0.39}}{\rec{+0.25}} & \cellms{\textbf{\rec{77.16}}}{\rec{0.32}}{\textcolor{improve}{\rec{+1.42}}} \\
\bottomrule
\end{tabular}
\label{tab:submetrics_hard}
\end{table*}

\section{Results on HDP}
\label{sec:supp_hdp}

HDP~\citep{zheng2026hdp} is the second planner to which we apply \methodname.
\Cref{tab:hdp} reports the six settings of both planners, each trained with non-reactive and with reactive traffic.
\methodname improves the score of HDP in every setting.

\section{Details of the experiments}
\label{sec:supp_ablation_details}

\Cref{tab:ablation_full} reports results in all six evaluation settings for every variant in \cref{sec:exp_takeaways}, evaluated at its selected checkpoint.

\paragraph{Selection of the checkpoint.}
A training-time evaluation scores Val14 R every 112 iterations in our simulator, and every row reports the last of the three consecutive checkpoints with the highest mean Val14 R gain over its runs, the rule that also selects the checkpoint of the reported runs.
The other five settings play no part in the selection.

\paragraph{Stalls.}
A stall is a failure in which the episode reaches the end of the scenario and the progress gate of the score is zero, which means that the vehicle covers less than 20\% of the expert's progress.
The pretrained planner has \rec{16} stalls on Test14-hard R.
A stall is released if the trained policy scores at least 50 in the scenario, and \cref{tab:events} reports the share of the released stalls, mean over six runs.

\section{Open-loop finetuning from logs}
\label{sec:supp_openloop}

The row ``Open-loop finetuning from logs'' of \cref{tab:main} applies open-loop finetuning to the Diffusion Planner with the online objective of DiffusionNFT~\citep{zheng2026diffusionnft}.
Each update takes 256 logged states from the training cache.
In each state, the planner samples one plan with the initial noise scale of 0.1 and 15 plans with the initial noise scale of 1.0.
Each of the 15 plans is sampled with offset guidance using the guidance mechanism of the Diffusion Planner~\citep{zheng2025diffusionplanner} that PlannerRFT~\citep{li2026plannerrft} also steers.
The offset guidance moves the predicted trajectory sideways by a random offset of up to 2\,m at every solver step.
The reward of a plan is its nuPlan score over the 4\,s horizon against the logged future.
The objective normalizes the rewards within each state and updates all weights of the network.
We train six runs for 30 updates each, until the score plateaus, and report the last checkpoint.

\begin{table}[b]
\centering
\footnotesize
\setlength{\tabcolsep}{3pt}
\caption{
\textbf{Estimated model FLOPs of RL training per run.}
Measured with the FLOP counter of PyTorch on one scene. A backward pass counts as two forward passes.
M, G, and P denote $10^{6}$, $10^{9}$, and $10^{15}$ floating-point operations (FLOPs).
\methodname: the settings of \cref{tab:hyper}.
PlannerRFT: its Table A1 (group size 8, 5 DDIM steps, 128 environments $\times$ 32 steps per iteration, GRPO mini-batch 4,096, 6 steps per epoch, 1 epoch).
One mini-batch sample is taken as one candidate trajectory with its five denoising steps, as in DiffusionDriveV2~\citep{zou2025diffusiondrivev2}.
The exploration policy of PlannerRFT, whose size is not stated, and its PPO updates are not counted, which favors PlannerRFT.
}
\begin{tabular}{@{}p{2.75in}r@{}}
\toprule
Measured cost & FLOPs \\
\midrule
Diffusion Planner, outside the denoising loop ($E$) & \rec{2.52}\,G \\
Diffusion Planner, one DDIM step of the DiT decoder ($D$) & \rec{0.117}\,G \\
Nudge agent, actor forward, one sample ($a$) & \rec{2.24}\,M \\
Nudge agent, one critic forward, one sample ($c$) & \rec{8.68}\,M \\
\midrule
Per environment step & GFLOPs \\
\midrule
\methodname: collection ($E + 5D + a$) & \rec{3.1} \\
\methodname: updates (25.6 sample-updates per step) & \rec{10.6} \\
\methodname: critic refits (three, amortized) & \rec{7.0} \\
\methodname: total & \rec{20.7} \\
PlannerRFT: collection ($E + 8 \times 5 \times D$) & \rec{7.2} \\
PlannerRFT: GRPO update, forward and backward per denoising step & \rec{10.5} \\
\midrule
Over the budget & PFLOPs \\
\midrule
\methodname, \rec{6.18}\,M steps & \rec{128} \\
PlannerRFT, 40\,M steps & \rec{710} \\
\bottomrule
\end{tabular}
\label{tab:flops}
\end{table}

\section{Sub-metrics on Test14-hard}
\label{sec:supp_submetrics}

\Cref{tab:submetrics_hard} breaks the Test14-hard R score into the devkit's terms and adds the score of the same runs on Test14-hard NR.
On Test14-hard R, open-loop finetuning from logs improves the progress-gate, progress, and comfort scores by 3.00, 3.08, and 1.22 points, respectively.
In contrast, its collision and time-to-collision scores fall by \rec{1.26} and \rec{1.28} points.
\methodname improves every term, most on the safety terms (collision \rec{+2.02}, time to collision \rec{+3.01}, drivable area \rec{+1.72}) and on the progress gate (\rec{+2.08}).
On Test14-hard NR, the gain of open-loop finetuning shrinks to \rec{+0.65}, within its spread over runs, while \methodname keeps \rec{+1.42}.

\section{Computational cost of RL training}
\label{sec:supp_flops}

PlannerRFT reports its hardware but not its training time, so we compare estimated model FLOPs per training run up to the reported checkpoint (\cref{tab:flops}).
We calculate training totals from measured forward-pass costs, counting each backward pass as two forward passes.
Under the accounting assumptions in \cref{tab:flops}, PlannerRFT's reported configuration requires approximately \rec{5.6} times as many model FLOPs as \methodname.
For \methodname, actor and critic updates, including critic refits, account for approximately \rec{85}\% of model FLOPs, and calls to the frozen planner account for approximately \rec{15}\%.


\setcounter{dbltopnumber}{3}
\begin{figure*}[p]
    \centering
    \includegraphics[width=0.9\linewidth]{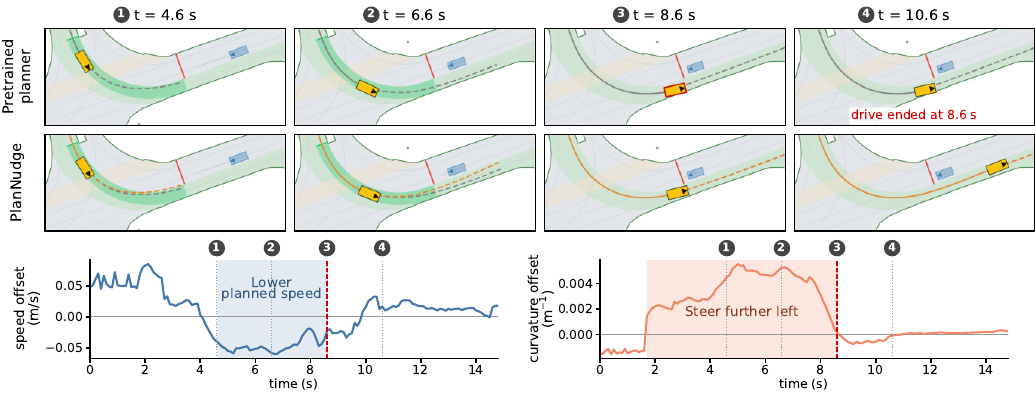}
    \caption{\textbf{A left turn, non-reactive traffic.} The pretrained planner does not turn sharply enough and leaves the drivable area at 8.6\,s. \methodname steers further left.}
    \label{fig:more_qualitative_offroad}
\end{figure*}

\begin{figure*}[p]
    \centering
    \includegraphics[width=0.9\linewidth]{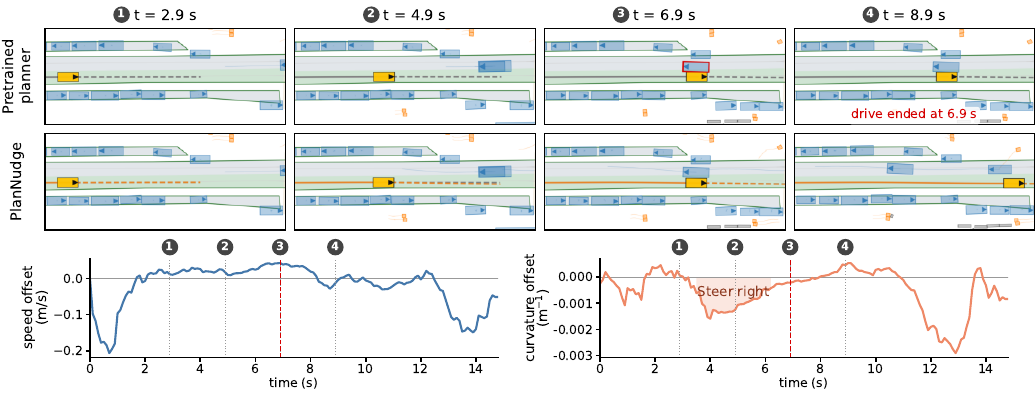}
    \caption{\textbf{A vehicle enters the ego vehicle’s lane from the left, non-reactive traffic.} \methodname steers right.}
    \label{fig:more_qualitative_cutin}
\end{figure*}

\begin{figure*}[p]
    \centering
    \includegraphics[width=0.9\linewidth]{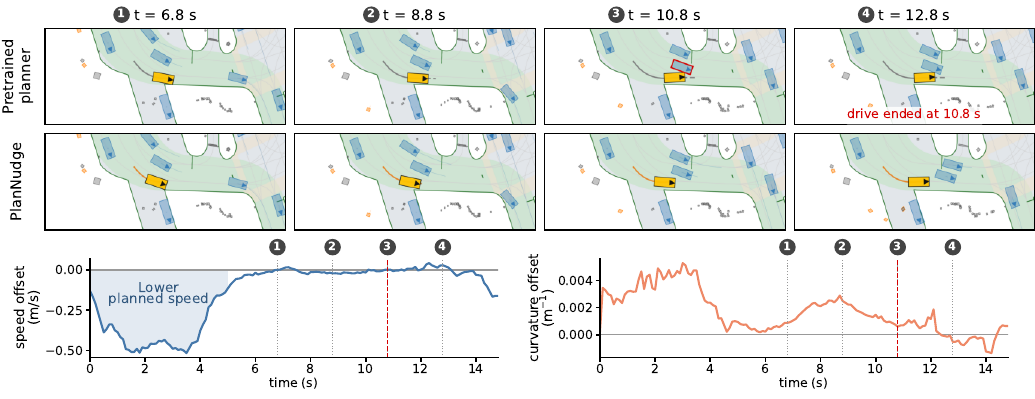}
    \caption{\textbf{A vehicle crosses ahead, non-reactive traffic.} \methodname slows down to let it pass.}
    \label{fig:more_qualitative_yield}
\end{figure*}


\begin{figure*}[p]
    \centering
    \includegraphics[width=0.9\linewidth]{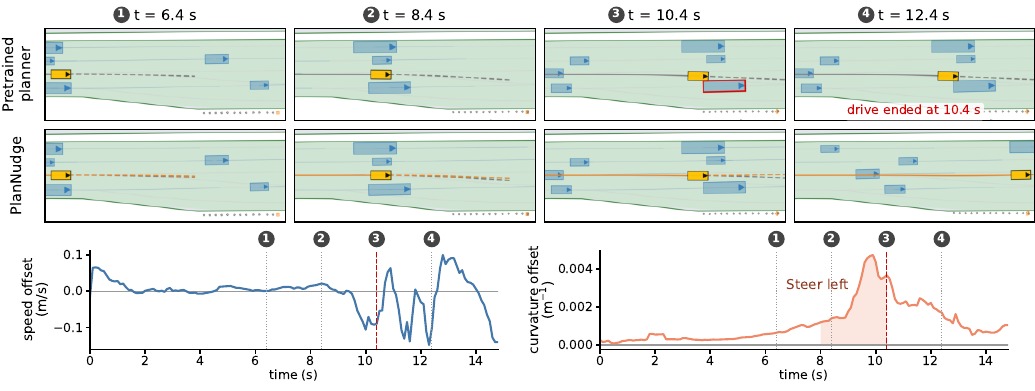}
    \caption{\textbf{A vehicle beside the ego in the right lane, reactive traffic.} The pretrained planner collides with it at 10.4\,s. \methodname steers left and passes it.}
    \label{fig:more_qualitative_r_rightlane}
\end{figure*}

\begin{figure*}[p]
    \centering
    \includegraphics[width=0.9\linewidth]{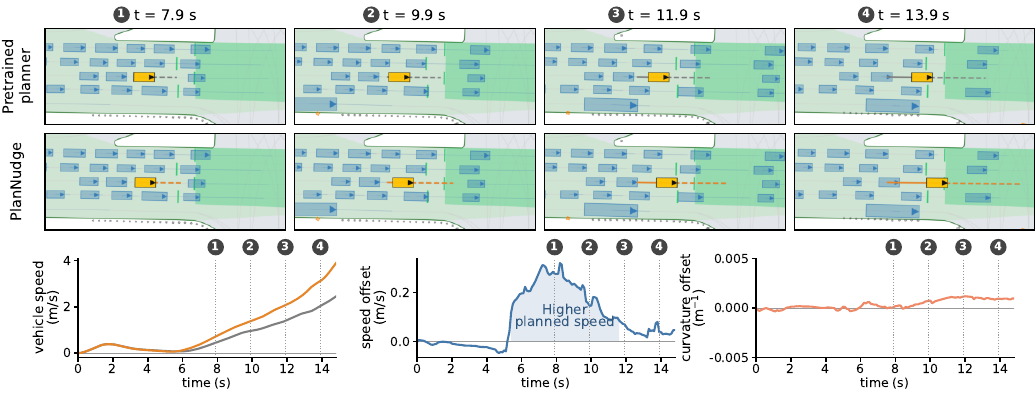}
    \caption{\textbf{A queue departs at an intersection, reactive traffic.} The pretrained planner follows too slowly. \methodname increases the planned speed. The left panel shows the vehicle speed in both drives, gray for the pretrained planner and orange for \methodname.}
    \label{fig:more_qualitative_r_queue}
\end{figure*}

\begin{figure*}[p]
    \centering
    \includegraphics[width=0.9\linewidth]{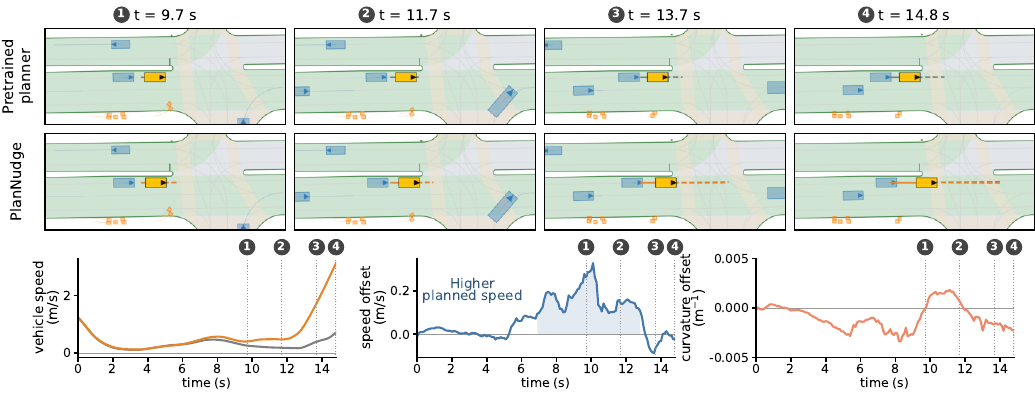}
    \caption{\textbf{Starting from a standstill, reactive traffic.} The pretrained planner barely moves until the end of the scenario. \methodname increases the planned speed and departs. The left panel shows the vehicle speed in both drives, gray for the pretrained planner and orange for \methodname.}
    \label{fig:more_qualitative_r_crossing}
\end{figure*}

\section{More qualitative results}
\label{sec:more_qualitative}

\Cref{fig:more_qualitative_offroad,fig:more_qualitative_cutin,fig:more_qualitative_yield} show three additional scenarios with non-reactive traffic, and \cref{fig:more_qualitative_r_rightlane,fig:more_qualitative_r_queue,fig:more_qualitative_r_crossing} show three with reactive traffic.
In each scenario, the pretrained planner fails and \methodname avoids failure.
All examples use the same training run and visualization conventions as \cref{fig:qualitative}.

\end{document}